\documentclass[sigplan,screen]{acmart}

\usepackage{mathtools}
\usepackage{graphicx}
\usepackage{textcomp}
\usepackage{multirow}
\usepackage{pifont}
\usepackage{xcolor}
\usepackage{diagbox}
\usepackage{tabularray}
\usepackage{url}
\usepackage{listings}
\usepackage{algorithm}
\usepackage[noend]{algpseudocode}

\usepackage{amssymb}
\usepackage{amsthm}

\usepackage{caption}
\usepackage{subcaption}
\usepackage{afterpage}
\usepackage[english]{babel}
\usepackage{amsthm} 
\usepackage[]{hyperref}
\usepackage{listings}
\usepackage{xcolor}
\definecolor{codegreen}{rgb}{0,0.6,0}
\definecolor{codegray}{rgb}{0.5,0.5,0.5}
\definecolor{codepurple}{rgb}{0.58,0,0.82}
\definecolor{backcolour}{rgb}{0.95,0.95,0.92}
\lstdefinestyle{mystyle}{
    backgroundcolor=\color{backcolour},   
    commentstyle=\color{codegreen},
    keywordstyle=\color{magenta},
    numberstyle=\tiny\color{codegray},
    stringstyle=\color{codepurple},
    basicstyle=\ttfamily\footnotesize,
    breakatwhitespace=false,         
    breaklines=true,                 
    captionpos=b,                    
    keepspaces=true,                 
    numbers=left,                    
    numbersep=5pt,                  
    showspaces=false,                
    showstringspaces=false,
    showtabs=false,                  
    tabsize=2
}
\newcommand\change[1]{#1}

\def \modelname {FSMoE}

\copyrightyear{2025}
\acmYear{2025}
\setcopyright{acmlicensed}\acmConference[ASPLOS '25]{Proceedings of the 30th ACM
International Conference on Architectural Support for Programming Languages and
Operating Systems, Volume 1}{March 30-April 3, 2025}{Rotterdam, Netherlands}
\acmBooktitle{Proceedings of the 30th ACM International Conference on Architectural
Support for Programming Languages and Operating Systems, Volume 1 (ASPLOS '25),
March 30-April 3, 2025, Rotterdam, Netherlands}
\acmDOI{10.1145/3669940.3707272}
\acmISBN{979-8-4007-0698-1/25/03}
\begin{document}

\title{ FSMoE: A Flexible and Scalable Training System
 for Sparse Mixture-of-Experts Models
  }
\author{Xinglin Pan}
\authornote{Equal contribution.}
\affiliation{%
  \institution{The Hong Kong University of Science and Technology (Guangzhou)}
  \city{Guangzhou}
  \country{China}
  }
  \email{xpan413@connect.hkust-gz.edu.cn}
\author{Wenxiang Lin}
\authornotemark[1]
\affiliation{%
  \institution{Harbin Institute of Technology, Shenzhen}
  \city{Shenzhen}
  \country{China}}
\email{wenxianglin@stu.hit.edu.cn}
\author{Lin Zhang}
\affiliation{%
  \institution{Hong Kong University of Science and Technology}
  \city{Hong Kong SAR}
  \country{China}}
\email{lzhangbv@connect.ust.hk}
\author{Shaohuai Shi}
\affiliation{%
  \institution{Harbin Institute of Technology, Shenzhen}
  \city{Shenzhen}
  \country{China}}  
\email{shaohuais@hit.edu.cn}
\author{Zhenheng Tang}
\affiliation{%
  \institution{The Hong Kong University of Science and Technology} \city{Hong Kong SAR}
  \country{China}}  
\email{zhtang.ml@ust.hk}
\author{Rui Wang}
\affiliation{%
  \institution{The Hong Kong University of Science and Technology (Guangzhou)}
  \city{Guangzhou}
  \country{China}}  
\email{rwang132@connect.hkust-gz.edu.cn}
\author{Bo Li}
\affiliation{%
  \institution{Hong Kong University of Science and Technology} \city{Hong Kong SAR}
  \country{China}}
\email{bli@cse.ust.hk}
\author{Xiaowen Chu}
\authornote{Also with Hong Kong University of Science and Technology, Hong Kong SAR.}
\affiliation{%
\institution{The Hong Kong University of Science and Technology (Guangzhou)}
\city{Guangzhou}
  \country{China}}
\email{xwchu@hkust-gz.edu.cn}

\renewcommand{\shortauthors}{Xinglin Pan et al.}

\begin{abstract}
Recent large language models (LLMs) have tended to leverage sparsity to reduce computations, employing the sparsely activated mixture-of-experts (MoE) technique. MoE introduces four modules, including token routing, token communication, expert computation, and expert parallelism, that impact model quality and training efficiency. To enable versatile usage of MoE models, we introduce FSMoE, a flexible training system optimizing task scheduling with three novel techniques: 1) Unified abstraction and online profiling of MoE modules for task scheduling across various MoE implementations. 2) Co-scheduling intra-node and inter-node communications with computations to minimize communication overheads. 3) To support near-optimal task scheduling, we design an adaptive gradient partitioning method for gradient aggregation and a schedule to adaptively pipeline communications and computations. We conduct extensive experiments with configured MoE layers and real-world MoE models on two GPU clusters. Experimental results show that 1) our \modelname{} supports four popular types of MoE routing functions and is more efficient than existing implementations (with up to a 1.42$\times$ speedup), and 2) \modelname{} outperforms the state-of-the-art MoE training systems (DeepSpeed-MoE and Tutel) by 1.18$\times$-1.22$\times$ on 1458 MoE layers and 1.19$\times$-3.01$\times$ on real-world MoE models based on GPT-2 and Mixtral using a popular routing function.
\end{abstract}


\ccsdesc[500]{Computing methodologies~Parallel algorithms}
\ccsdesc[500]{Computing methodologies~Machine learning}

\keywords{Distributed Deep Learning; Large Language Model; Mixture-of-Experts; Training System; Scheduling}

\maketitle 

\section{Introduction}

In recent years, the concept of sparsely-activated Mixture-of-Experts (MoE) layers has garnered considerable attention~\cite{DBLP:Chowdhery2022plam, DBLP:danny2023palme, DBLP:Shazeer2017outrageously, DBLP:Lepikhin2021gshard,jiang2024mixtral, deepseekv2} in large language models (LLMs) as MoE can scale up the model size while keeping the computational cost for training be sub-linearly increased. MoE models incorporate multiple experts in MoE layers, and each expert represents a specialized feed-forward network~(ffn) trained for specific subtasks (with particular input tokens). These MoE layers utilize a gating mechanism using a routing function (e.g., employing softmax activation~\cite{DBLP:Lepikhin2021gshard}) to dynamically assign data samples to the appropriate experts. This approach allows for scaling up model sizes. For instance, the Switch Transformer~\cite{DBLP:Fedus2022switch} scales up to 1.5 trillion parameters with 15 MoE layers, each consisting of 2048 experts, surpassing its dense model (w/o MoE) that has only several billion parameters. The MoE technique has significantly improved model performance in domains such as natural language processing~\cite{zuo2021taming, zhou2022mixture}, computer vision~\cite{huang2023experts}, speech recognition~\cite{you2021speechmoe}, and recommendation systems~\cite{ma2018modeling}. 

Though the MoE technique has achieved remarkable success in many AI models, it is still actively studied in both algorithms and systems. From the algorithm's point of view, how to train the experts effectively and efficiently with what input tokens is still an open problem, which means the routing function plays an important role in the model generalization capability~\cite{DBLP:Shazeer2017outrageously,DBLP:Lepikhin2021gshard,lewis2021base,zhou2022mixture,DBLP:Fedus2022switch,chi2022representation,puigcerver2023sparse}. From the system's point of view, dedicated MoE systems (e.g., FastMoE~\cite{he2021fastmoe}, DeepSpeed-MoE~\cite{DBLP:Rajbhandari22deepspeedmoe}, FlexMoE~\cite{nie2023flexmoe}, Tutel~\cite{hwang2023tutel}, and ScheMoE~\cite{shi2024schemoe}) are designed to support the study of algorithms and deployment of MoE models. Atop of these systems, there exist many optimization strategies to improve MoE training efficiency by designing efficient communication collectives (e.g., AlltoAll algorithms~\cite{DBLP:Rajbhandari22deepspeedmoe,hwang2023tutel,aminabadi2022deepspeed,DBLP:ma2022bagualu}), optimizing sparsity computation~\cite{zheng2023pit}, and scheduling communication and computation tasks~\cite{hwang2023tutel,he2022fastermoe,shi2023pipemoe,li2023lina,zhai2023smartmoe,liu2023janus,pan2024parm, shi2024schemoe, chen2024centauri, jiang2024lancet} to enable the overlap of communication tasks with computing tasks. \textit{However, these systems have two main limitations: 1) they only support limited routing functions and are inflexible to support newly introduced routing mechanisms, and 2) they are optimized with particular parallelisms and are sub-optimal in some commonly used scenarios.}

As the scale of LLM continues to expand, various parallel paradigms have emerged to address the escalating computational demands. In addition to the three well-established parallel paradigms for traditional models, namely Data Parallelism (DP)~\cite{DBLP:dean2012large}, Model Parallelism (MP)~\cite{DBLP:dean2012large}, and Pipeline Parallelism (PP)~\cite{DBLP:huang2019gpipe}, MoE models introduce two additional parallel paradigms: Expert Parallelism (EP)~\cite{DBLP:Shazeer2017outrageously} and Expert-Sharding Parallelism (ESP)~\cite{DBLP:Rajbhandari22deepspeedmoe} to enable large experts to be trained on GPU clusters (\S\ref{subsec:parallelism}). These parallelisms significantly impact the system scalability by the communication time incurred by the various parallel paradigms. Research studies such as~\cite{DBLP:Lepikhin2021gshard, hwang2023tutel, DBLP:liu2022gating, li2023lina} have reported that the AlltoAll communication time comprises a substantial portion, ranging from 30\% to 60\%, of the overall time required for executing the MoE layers on high-end GPU or TPU clusters. This issue becomes more severe when DP, MP, EP, and ESP (DP+MP+EP+ESP) are employed simultaneously for large-scale training~\cite{hwang2023tutel}. Consequently, rapid changes in MoE routing mechanisms and complicated parallelisms present a significant modularisation difficulty to flexibly support newly designed MoE components and a performance challenge to optimally schedule different time-consuming tasks.

To this end, in this work, we propose FSMoE~\footnote{Code Repository: \url{https://github.com/xpan413/FSMoE}.}, a flexible and efficient MoE system with near-optimal task scheduling, to efficiently train MoE models. First, to enhance extensibility, we design unified abstraction and online profiling of MoE modules for task scheduling across various MoE implementations. Second, to better schedule the communication and computing tasks in DP+MP+EP+ESP, we analyse the possibility of overlapping the inter-node and intra-node communications and propose an efficient schedule to pipeline inter-node and intra-node communication tasks as well as computation tasks in a common scenario where the MP and ESP group is configured to align with the number of GPUs within a node. Third, to support near-optimal task scheduling, we design a gradient partitioning method for fully overlapping the gradient aggregation with other tasks. Our main technical contributions are summarized as follows.
\begin{itemize}
  \item We modularize all possible operations in an MoE layer to support various MoE components, including gating function, data layout, collective communication, expert computation, etc. We have provided the implementations of four popular types of gating functions in our system, which are more efficient than their original implementations.
  \item Based on the modularized operations, we propose an adaptive optimal scheduling algorithm to pipeline both intra-node and inter-node communication tasks as well as computing tasks to improve the training efficiency in common scenarios where the group of MP and ESP matches the number of GPUs per node.
  \item We design an adaptive gradient partitioning method to hide the communication cost of the gradient aggregation by pipelining communications with computations and avoid the contention between different inter-node communications.
  \item We conduct extensive experiments on a 48-GPU cluster and a 32-GPU cluster using customized MoE layers and real-world MoE models. Experimental results show that: (1) FSMoE outperforms Tutel~\cite{hwang2023tutel} (with its improved algorithm PipeMoE~\cite{shi2023pipemoe}) by $1.18\times$ to $1.22\times$ on training 1458 different configured cases. (2) FSMoE runs $1.19\times$ to $3.01\times$ faster on average than the state-of-the-art MoE systems (Tutel and DeepSpeed-MoE~\cite{DBLP:Rajbhandari22deepspeedmoe}) on training two real-world MoE models based on GPT-2 and Mixtral.
\end{itemize}
\begin{table}[!t]
	\centering
	\caption{Notations.}
	\label{tab:notations}
	\begin{tabular}{|l|l|}
		\hline
		Name &  Description \\\cline{1-2}
		\hline
		\hline
        $P$ & \# of GPUs \\
        $r$ & \# of the pipeline degree \\
		$B$ & \# of samples per GPU (or local mini-batch size) \\
		$L$ & \# of tokens per sample (or sequence length) \\
		$E$ & total number of experts \\
        $k$ & top-$k$ experts should be selected for each token \\
        $f$ & factor to control expert's maximum token count \\
        
		$M$ & embedding size of a token \\
		$H$ & hidden size of the feed-forward layer in experts \\
        $N_{head}$& \# of heads in the attention layer \\
        $N_{DP}$ & \# of workers in each DP group \\
		$N_{MP}$ & \# of workers in each MP group \\
        $N_{EP}$ & \# of workers in each EP group \\
        $N_{ESP}$ & \# of workers in each ESP group \\
        $N_{PP}$ & \# of workers in each PP group \\
          \hline
	\end{tabular}
\end{table}

\section{Background and Motivations}
For ease of presentation, we provide a summary of the essential notations employed in the paper, presented in Table~\ref{tab:notations}.

\subsection{Mixture-of-Experts Layer}
\begin{figure}[!t]
	\centering
\includegraphics[width=\linewidth]{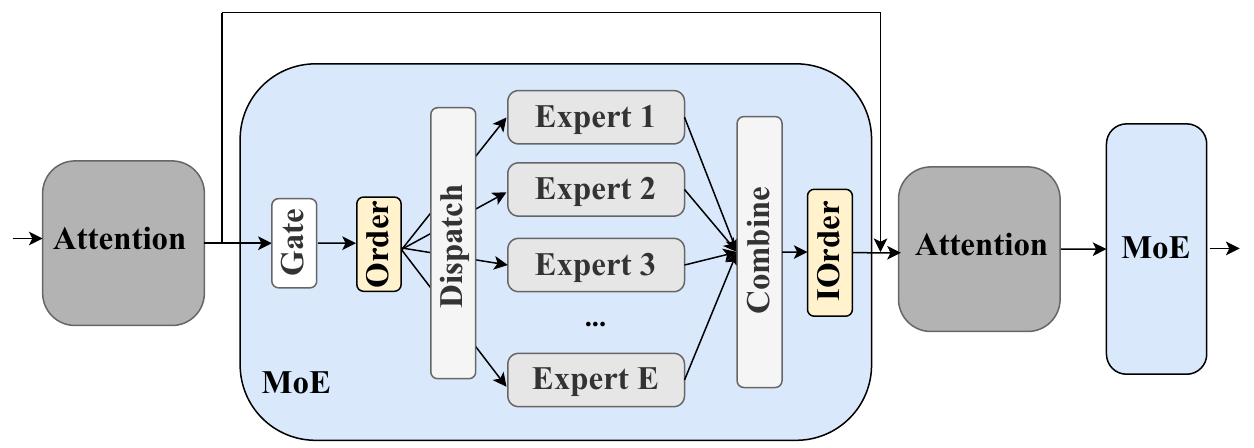}
	\caption{A typical MoE structure with $E$ experts.}
 \label{fig:moe-layer}
\end{figure}

In modern MoE models, which are typically built atop the Transformer~\cite{vaswani2017attention} architecture, an MoE layer is used to replace the ffn layer. As shown in Fig.~\ref{fig:moe-layer}, the MoE layer comprises three core components: a gating function, an ordering function (and its reverse operation, i.e., the I-ordering function) and a set of $E$ experts. 


\textbf{Gating Function.} The gating function plays a pivotal role in assigning tokens to specific experts. During each training iteration, the input data (denoted as $I$) of the MoE layer has a shape of ($B, L, M$), where $B$ represents the mini-batch size, $L$ represents the sequence length per sample, and $M$ represents the embedding size. To determine the activation of experts, $I$ is divided into multiple parts based on the gating function.

GShard~\cite{DBLP:Lepikhin2021gshard} employs a noisy Top-k Gate, denoted as $G(I)=\operatorname{Softmax}(\operatorname{KeepTopK}(H(I), k)),
$
where $H(I)$ adds noises to the input $I$ through a specific transformation:
$$
\begin{gathered}
H(I)_i=\left(I \cdot W_g\right)_i+\mathcal{N}(0, 1) \cdot \operatorname{Softplus}\left(\left(I \cdot W_{\text {noise }}\right)_i\right),
\end{gathered}
$$
and the function 
$\text{KeepTopK}(v, k)$ retains the top $k$ values of a vector $v$, setting the rest to negative infinity:
$$
\begin{gathered}
\text { KeepTopK }(v, k)_i= \begin{cases}v_i & \text { if } v_i \text { is in the top } k \text { values of } v . \\
-\infty & \text { otherwise. }\end{cases}
\end{gathered}
$$
In $\text{KeepTopK}(v, k)$, \(W_g\) and \(W_{\text{noise}}\) are two trainable weights.

In BASE~\cite{lewis2021base} and StableMoE~\cite{zhai2023smartmoe} models, the sigmoid gate is employed, defined by \( H(I)_i = (I \cdot W_g)_i \). The output from the expert is scaled by \(\sigma(H(I)_i)\). If this output contributes positively to \( I \), optimizing the training goal (such as minimizing cross-entropy loss in language modelling) increases the gate value, favouring the selection of the same expert. In X-MoE~\cite{chi2022representation}, a low-rank linear projection \(W_{proj} I\) is employed to segregate the direct interaction between the hidden vector \(I\) and the expert embedding \(W_g\). This approach effectively mitigates the issue of cascaded collapse in representations. Subsequently, these representations undergo an l2 normalization process to be scaled appropriately. The formula can be expressed as follows: $s_i=\operatorname{cos}(W_{proj} I, W_g)$. An expert choice method~\cite{zhou2022mixture} independently selects top-$k$ tokens
for each expert, denoted as $ G(I)=\operatorname{Softmax}(\operatorname{KeepTopK}((I \cdot W_g)^\intercal, k))$.

The effectiveness of gating functions is assessed using specific models and datasets. 
For example, EC~\cite{zhou2022mixture} is evaluated through casual language modelling tasks, whereas X-MoE~\cite{chi2022representation} is assessed via masked language modelling tasks. 
When encountering new challenges, developers cannot determine the most suitable gating functions for the task without conducting practical tests. Therefore, incorporating a diverse range of gating functions enhances the robustness for developers.

\textbf{Ordering and I-Ordering Functions.} The ordering function transforms the input tensor layout before dispatched. Typically, the format changes from $(B, L, M)$ to $(E, T, M)$, where $T$ denotes the maximum tokens per expert. $T$ is determined using the formula $T \coloneqq k\times f\times B\times L / E$, where $f$ is a control factor. Each row of $G$ (i.e., $G[i,:,:]$) aligns with the data for the $i\text{-th}$ expert ($i$ ranges from 1 to $E$). There are two main types of ordering functions: 1) GShard~\cite{DBLP:Lepikhin2021gshard} ordering, which uses a combination of einsum and matrix multiplication, and 2) Tutel~\cite{hwang2023tutel} ordering, which employs SIMT-efficient sparse operations. The I-ordering function serves as a reverse function of the ordering function, allowing for the data layout to be adjusted back to its original form.


\textbf{Experts.} Typically, each expert in the MoE layer is a compact neural network consisting of several feed-forward layers followed by an activation function~\cite{DBLP:Lepikhin2021gshard,jiang2024mixtral}. Take a two-layer expert as an example, the first layer has a weight matrix with a shape of ($M, H$), while the second layer has a shape of ($H, M$), where $H$ represents the size of the hidden layer so that the output of expert has the shape with the input. For an MoE layer with $E$ experts, we denote the $i$-th expert as $e_i$, and its output can be denoted as $Q_i = e_i(G[i,:,:])$.
To obtain the final output of the MoE layer, the outputs of all experts are combined into a unified tensor $[Q_1, Q_2, ..., Q_{E}]$. This tensor can be reshaped as ($B, L, M$) to facilitate further processing in subsequent layers.
    
Despite the expansion in the model size in MoE models, the increase in their computational cost is marginal. However, the size of these models has grown to such an extent that they cannot be loaded into the memory of a single device. As a result, distributed training becomes essential for training MoE models, leveraging multiple devices to handle the computational and memory demands, which easily introduces significant communication overheads. A benchmark of the training time breakdown with two popular MoE models is conducted on our 32-GPU and 48-GPU testbeds (details in \S\ref{sec:evaluation}) is shown in Table~\ref{tab:breakdown}. It demonstrates that communication overhead typically contributes over 50\% to the overall training, indicating the necessity of optimizing communication performance.


\begin{table*}[h]
\centering
	\caption{Time performance (iteration time in millisecond) of each operation in a transformer layer of two real-world models, GPT2-XL~\cite{radford2019language} and Mixtral7B~\cite{jiang2024mixtral}, with $B=4$ and $L=1024$ for two testbeds in Table~\ref{tab:server-config}. The numbers in the brackets represent each operation's portion of the forward and backward time.}
 \resizebox{\textwidth}{!}{%
\begin{tabular}{llllllllll}
\hline
\multicolumn{2}{|l|}{\multirow{2}{*}{Testbeds/Breakdown}}                        & \multicolumn{4}{c|}{Communication}                                                                                                                 & \multicolumn{4}{c|}{Computation}                                                                                                             \\ \cline{3-10} 
\multicolumn{2}{|l|}{}                                                           & \multicolumn{1}{l|}{AlltoAll}      & \multicolumn{1}{l|}{AllReduce}      & \multicolumn{1}{l|}{AllGather}     & \multicolumn{1}{l|}{ReduceScatter} & \multicolumn{1}{l|}{Experts}       & \multicolumn{1}{l|}{Routing}     & \multicolumn{1}{l|}{Order}       & \multicolumn{1}{l|}{Attention}    \\ \hline
\multicolumn{1}{|l|}{\multirow{4}{*}{A}} & \multicolumn{1}{l|}{GPT2-Forward}     & \multicolumn{1}{l|}{6.9(31.16\%)}  & \multicolumn{1}{l|}{0(0\%)}         & \multicolumn{1}{l|}{4.6(20.83\%)}  & \multicolumn{1}{l|}{5.4(24.46\%)}  & \multicolumn{1}{l|}{3.1(14.04\%)}  & \multicolumn{1}{l|}{0.1(0.45\%)} & \multicolumn{1}{l|}{0.3(1.36\%)} & \multicolumn{1}{l|}{1.7(7.7\%)}   \\ \cline{2-10} 
\multicolumn{1}{|l|}{}                   & \multicolumn{1}{l|}{GPT2-Backward}    & \multicolumn{1}{l|}{6.9(21.27\%)}  & \multicolumn{1}{l|}{5.26(16.26\%)}  & \multicolumn{1}{l|}{4.6(14.22\%)}  & \multicolumn{1}{l|}{5.4(16.7\%)}   & \multicolumn{1}{l|}{6.1(18.86\%)}  & \multicolumn{1}{l|}{0.1(0.31\%)} & \multicolumn{1}{l|}{0.4(1.24\%)} & \multicolumn{1}{l|}{3.6(11.13\%)} \\ \cline{2-10} 
\multicolumn{1}{|l|}{}                   & \multicolumn{1}{l|}{Mixtral-Forward}  & \multicolumn{1}{l|}{19.5(29.8\%)}  & \multicolumn{1}{l|}{0(0\%)}         & \multicolumn{1}{l|}{12.3(18.73\%)} & \multicolumn{1}{l|}{13.7(20.86\%)} & \multicolumn{1}{l|}{15.6(23.76\%)} & \multicolumn{1}{l|}{0.1(0.15\%)} & \multicolumn{1}{l|}{0.3(0.46\%)} & \multicolumn{1}{l|}{4.1(6.24\%)}  \\ \cline{2-10} 
\multicolumn{1}{|l|}{}                   & \multicolumn{1}{l|}{Mixtral-Backward} & \multicolumn{1}{l|}{19.6(17.45\%)} & \multicolumn{1}{l|}{26.45(23.59\%)} & \multicolumn{1}{l|}{12.3(10.97\%)} & \multicolumn{1}{l|}{13.7(12.22\%)} & \multicolumn{1}{l|}{31.8(28.36\%)} & \multicolumn{1}{l|}{0.1(0.09\%)} & \multicolumn{1}{l|}{0.5(0.45\%)} & \multicolumn{1}{l|}{7.7(6.87\%)}  \\ \hline
\multicolumn{1}{|l|}{\multirow{4}{*}{B}} & \multicolumn{1}{l|}{GPT2-Forward}     & \multicolumn{1}{l|}{11.2(20.7\%)}  & \multicolumn{1}{l|}{0(0.0\%)}       & \multicolumn{1}{l|}{15.5(28.7\%)}  & \multicolumn{1}{l|}{15.7(29.1\%)}  & \multicolumn{1}{l|}{6.7(12.4\%)}   & \multicolumn{1}{l|}{0.1(0.2\%)}  & \multicolumn{1}{l|}{0.3(0.6\%)}  & \multicolumn{1}{l|}{4.5(8.3\%)}   \\ \cline{2-10} 
\multicolumn{1}{|l|}{}                   & \multicolumn{1}{l|}{GPT2-Backward}    & \multicolumn{1}{l|}{11.2(15.7\%)}  & \multicolumn{1}{l|}{7.3(10.3\%)}    & \multicolumn{1}{l|}{15.5(21.8\%)}  & \multicolumn{1}{l|}{15.2(21.3\%)}  & \multicolumn{1}{l|}{13(18.3\%)}    & \multicolumn{1}{l|}{0.1(0.1\%)}  & \multicolumn{1}{l|}{0.3(0.4\%)}  & \multicolumn{1}{l|}{8.6(12.1\%)}  \\ \cline{2-10} 
\multicolumn{1}{|l|}{}                   & \multicolumn{1}{l|}{Mixtral-Forward}  & \multicolumn{1}{l|}{28.3(15.9\%)}  & \multicolumn{1}{l|}{0.0(0.0\%)}     & \multicolumn{1}{l|}{39.6(22.3\%)}  & \multicolumn{1}{l|}{40.8(23.0\%)}  & \multicolumn{1}{l|}{58.5(33.0\%)}  & \multicolumn{1}{l|}{0.1(0.1\%)}  & \multicolumn{1}{l|}{0.7(0.4\%)}  & \multicolumn{1}{l|}{9.5(5.4\%)}   \\ \cline{2-10} 
\multicolumn{1}{|l|}{}                   & \multicolumn{1}{l|}{Mixtral-Backward} & \multicolumn{1}{l|}{30.8(10.8\%)}  & \multicolumn{1}{l|}{32.1(11.3\%)}   & \multicolumn{1}{l|}{40.1(14.1\%)}  & \multicolumn{1}{l|}{41.8(14.7\%)}  & \multicolumn{1}{l|}{119.7(42.1\%)} & \multicolumn{1}{l|}{0.2(0.1\%)}  & \multicolumn{1}{l|}{1.2(0.4\%)}  & \multicolumn{1}{l|}{18.1(6.4\%)}  \\ \hline
\end{tabular}
}
\label{tab:breakdown}
\end{table*}

\subsection{Paradigms of Parallelism}\label{subsec:parallelism}
The hybrid parallelism with DP, MP, EP, and ESP is required to train large-scale MoE models on a GPU cluster. 

\textbf{Data Parallelism.} In distributed DL, the data parallelism (DP) training technique has become a de-facto method~\cite{DBLP:dean2012large,jia2018highly,You2020Large}, where a mini-batch of samples is distributed to the workers in the DP group. During backpropagation, the gradients of each worker in the same DP group are aggregated through an AllReduce operation (we call Gradient-AllReduce afterwards) so that they can use the identical gradient to update model parameters.

\textbf{Model Parallelism.} Model Parallelism (MP)~\cite{DBLP:dean2012large,narayanan2021efficient} is a technique that divides model parameters among multiple workers to facilitate parallel computation. Each worker performs its computations independently, and subsequently, the outputs from all workers are combined through an AllReduce collective operation. Notably, when the MP group is configured as the number of GPUs within a node, which is very common, the communication involved in MP is considered intra-node communication, while the collective communication for gradient aggregation involves inter-node communication.

\textbf{Expert Parallelism.}
In Expert Parallelism (EP)~\cite{DBLP:Shazeer2017outrageously,DBLP:Lepikhin2021gshard}, experts are assigned to different GPUs, ensuring that each device handles a specific subset of experts. After the data passes through the gating function, the rows of the tensor $G$ ($G[i,:,:]$) on each device correspond to the data assigned to the respective $i$-th expert ($i=1, 2, \cdots, E$). As the experts are distributed across multiple devices, the dispatch operation uses a collective communication technique called AlltoAll Dispatch. This approach facilitates sending tokens to their respective experts for computation. Subsequently, the outputs generated by all experts are combined using another AlltoAll operation, known as AlltoAll Combine, for further processing.

\textbf{Expert-Sharding Parallelism.} 
When training large-scale MoE models, the number of workers $P$ may exceed the number of experts $E$. In such cases, expert-sharding parallelism (ESP)~\cite{hwang2023tutel,DBLP:Rajbhandari22deepspeedmoe,singh2023hybrid} can be employed to distribute the workload evenly across all workers. ESP groups are formed to uniformly partition the experts among the GPUs within each group, similar to MP. This enables parallel computation of expert outputs across all workers within the ESP group.

The combination of EP and ESP is required to place each MoE layer across multiple GPUs, which introduces additional communication operators~\cite{singh2023hybrid,pan2024parm}, namely ESP-AllGather and ESP-ReduceScatter. ESP-AllGather ensures that the input data is uniformly distributed among all workers within the ESP group, while ESP-ReduceScatter is used to aggregate the outputs of expert shards within the ESP group and split them back into the original structure of the input. The number of GPUs in an ESP group is denoted as $N_\text{ESP}$.
Notably, when the ESP group is configured to align with GPUs within a node, the ESP-AllGather and ESP-ReduceScatter operations involve intra-node communication while the AlltoAll operation introduced by EP entails inter-node communication, enabling the overlaps between ESP-AllGather/ESP-ReduceScatter and AlltoAll. In this work, we mainly discuss the schedule under this case.

\begin{figure*}[!ht]
	\centering
\includegraphics[width=\linewidth]{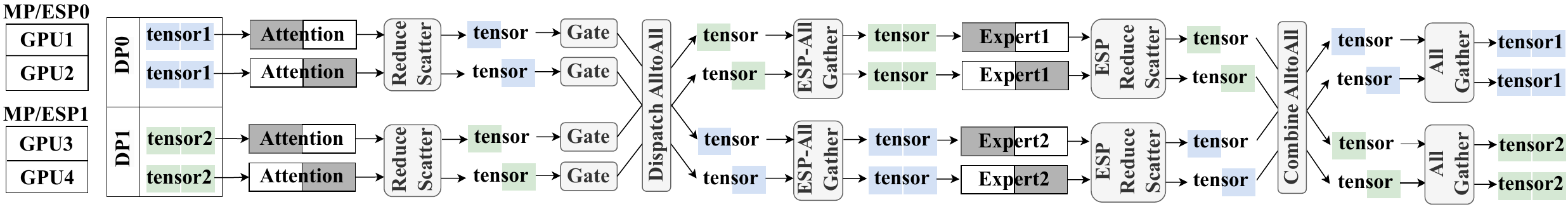}
	\caption{An example of $N_{\text{DP}}=N_{\text{MP}}=N_{\text{EP}}=N_{\text{ESP}}=2$. The attention is partitioned into two parts across MP groups, and the two experts are distributed to the two EP groups (GPU1 and GPU3, as well as GPU2 and GPU4) in EP, and each expert is further partitioned into two shards across the ESP group. The blue and green rectangles indicate the data tensors.}
 \label{fig:full_example}
\end{figure*}

An example of training an MoE model~\cite{singh2023hybrid} with DP, MP, EP, and ESP is shown in Fig.~\ref{fig:full_example}, where $N_{\text{DP}}=N_{\text{MP}}=N_{\text{EP}}=N_{\text{ESP}}=2$. In this example, two different tensors (or two mini-batches of samples) from \textit{the DP group} go through the attention layer partitioned across \textit{two MP groups} and are divided into half by using a ReduceScatter operation introduced by MP. Then two split tensors find selected experts partitioned across \textit{two ESP groups} by the gating function and are dispatched into the corresponding devices across \textit{two EP groups} (GPU1 and GPU3; GPU2 and GPU4) through an AlltoAll operation. Before the expert computation, split tensors should be combined through an AllGather collective across the two ESP groups called ESP-AllGather. Then, after the experts computation, tensors are divided into half again by another ReduceScatter operation introduced by ESP, which is called ESP-ReduceScatter, and they are sent back to their original workers through another AlltoAll operation. Finally, another AllGather operation is performed for these tensors across the MP groups to finalize the output. It is seen that it requires several key components and complicated parallelisms to train MoE models, which motivates our designed system to provide a flexible and scalable MoE training system.

\subsection{Motivations}\label{schechanllenge}

\textbf{A Flexible MoE framework.}
A flexible MoE framework should efficiently combine different routing functions~\cite{DBLP:Shazeer2017outrageously,DBLP:Lepikhin2021gshard,lewis2021base,zhou2022mixture,DBLP:Fedus2022switch,chi2022representation,puigcerver2023sparse}, order functions~\cite{hwang2023tutel, he2022fastermoe}, expert blocks~\cite{DBLP:Brown2020GPT3, jiang2024mixtral}, and AlltoAll algorithms~\cite{DBLP:Rajbhandari22deepspeedmoe,hwang2023tutel,aminabadi2022deepspeed,DBLP:ma2022bagualu}. This integration should be achieved with minimal complex programming for additional customization. The aim is to comprehensively address all types of overlaps, like communication with communication or computing, particularly when dealing with diverse parallel groups like integrating DP, MP, EP, and ESP~(\textsection\ref{sec:system_design}).

\textbf{Optimizing Network Communication.} As shown in Fig.~\ref{fig:nooverlap}, various parallel paradigms (e.g., DP, MP, EP, ESP) comprise a substantial portion of the overall iteration time. To mitigate the communication cost associated with the MoE layer, prior research (e.g., Tutel~\cite{hwang2023tutel}, PipeMoE~\cite{shi2023pipemoe}, FasterMoE~\cite{he2022fastermoe}) has explored overlapping AlltoAll with experts as illustrated in Fig.~\ref{fig:pipemoe}. However, they do not explore the overlapping ESP-AllGather/ESP-ReduceScatter (intra-node communication) with AlltoAll Dispatch/Combine (inter-node communication), diminishing network efficiency. This motivates us to pipeline inter-node and intra-node communication as shown in Fig.~\ref{fig:ourswopg} (\textsection\ref{sec4}).

\textbf{Optimizing Forward and Backward Separately.} Existing systems (e.g., Tutel~\cite{hwang2023tutel} and DeepSpeed-MoE~\cite{DBLP:Rajbhandari22deepspeedmoe}) typically use the same pipeline degree (i.e., the number of split input chunks for the overlaps) for both forward and backward propagation during training. However, the ideal degree may vary between these two phases due to their distinct computational requirements. For example, backward propagation involves additional computations to calculate the gradient of weights. Our extensive experiments on 1,458 MoE configurations (details in Table~\ref{tab:moe-configs}) reveal that 912 cases exhibit varied optimal pipeline degrees, tested on a 32-GPU cluster with 8 nodes (details in Table~\ref{tab:server-config}). Therefore, adaptively determining the pipeline degrees for both forward and backward phases is needed to achieve better training efficiency (\textsection\ref{sec4.4}).

\textbf{Co-Design in Backward Propagation and Gradient Synchronization.} 
Since Gradient-AllReduce (introduced by the weight synchronization in DP) and AlltoAll are both inter-node communication, Gradient-AllReduce can not be directly overlapped with the whole MoE layer as shown in Fig.~\ref{fig:pipemoe} and Fig.~\ref{fig:ourswopg} which only overlap Gradient-AllReduce with non-MoE parts. Consequently, designing overlaps for Gradient-AllReduce without considering MoE layers tends to result in sub-optimal solutions. A co-design that considers the AlltoAll operation and adjusts the partitioning of gradients for optimal overlapping remains unexplored (\textsection\ref{sec5}).



\section{FSMoE: System Design}\label{sec:system_design}
We propose FSMoE, a flexible and scalable MoE framework for distributed training. Our framework has three main characteristics: 1) modularization and non-invasive modification, 2) isolation of front-end API definition and back-end task scheduling, and 3) easy schedule of different tasks.

\subsection{Modularization and Non-Invasive Modification}
In our \modelname{} framework, the MoE layer is divided into six distinct sub-modules, namely: \textit{Gate}, \textit{Order}, \textit{I-Order}, \textit{Dispatch}, \textit{Combine}, \textit{Expert}.

\textit{Gate:} The \textit{Gate} sub-module determines how tokens are assigned to different experts for calculation. We pre-implement four routing functions: GShard routing~\cite{DBLP:Lepikhin2021gshard}, Sigmoid~\cite{lewis2021base, dai2022stablemoe} routing, X-MoE routing~\cite{chi2022representation}, and SoftMoE routing~\cite{puigcerver2023sparse}.

\textit{Order \& I-Order:} The \textit{Order} sub-module transforms the input tensor layout before it is dispatched. Typically, the format changes from $(B, L, M)$ to $(E, T, M)$. We pre-implement two types of ordering functions: 1) GShard~\cite{DBLP:Lepikhin2021gshard} ordering, which uses a combination of einsum and matrix multiplication, and 2) Tutel~\cite{hwang2023tutel} ordering, which employs SIMT-efficient sparse operations. The \textit{I-Order} sub-module serves as a reverse operation of the \textit{Order} sub-module, allowing for the data layout to be adjusted back to its original form.

\textit{Dispatch \& Combine:}
The \textit{Dispatch} sub-module handles the collective communication for the token-to-expert dispatch. It allows users to customize the collective communication algorithm without impacting our scheduler. To facilitate this customization, we pre-implement the default A2A algorithm provided by NCCL (NCCL-A2A)~\cite{all2allnccl2022}, 1DH-A2A proposed
by Hetu~\cite{DBLP:nie2022hetumoe}, 2DH-A2A proposed by Tutel~\cite{hwang2023tutel} and DeepSpeed-MoE~\cite{DBLP:Rajbhandari22deepspeedmoe}. This customization ensures optimal dispatching based on user-specific needs. The \textit{Combine} sub-module serves as a reverse operation of the \textit{Dispatch} sub-module.


\textit{Expert:} The Expert sub-module manages the computation task. Modules derived from ``torch.nn.Module'' can serve as the expert component. We offer two variants of these networks: the GPT feed-forward network~\cite{DBLP:Brown2020GPT3} and the Mixtral feed-forward network~\cite{jiang2024mixtral}.

\textit{Hooks:} In our framework, we offer a range of hooks for non-intrusive modification, including \textit{BeforeMoeStartHook}, \textit{BeforeDispatchHook}, \textit{AfterDispatchHook}, \textit{BeforeCombineHook}, and \textit{AfterCombineHook}, as well as \textit{BeforeMoeEndHook}. These hooks facilitate various adjustments without requiring invasive changes. For example, in handling multimodal data, \textit{BeforeMoeStartHook} and \textit{BeforeMoeEndHook} can be utilized to reformat inputs to conform to the standard MoE layer configuration. In another scenario, such as communication compression, \textit{BeforeDispatchHook} is used to compress the tensor before dispatch, while \textit{AfterDispatchHook} serves to decompress it afterward, ensuring efficient extension without the need for fundamental code modifications.
\subsection{Generic Scheduler}
The FSMoE framework boasts a versatile scheduling capability for task pipelines, independent of specific API definitions. It includes a profiler for evaluating the time efficiency of various tasks. Utilizing data collection and predictive modeling, it strategically coordinates sub-modules in MoE, achieving higher efficiency. It includes two main parts: \textit{front-end} and \textit{back-end}.

\textbf{\textit{Front-end}}. Developers select or build their MoE layers. A profiling model then evaluates the execution time of API definitions across various input sizes, using machine learning algorithms (e.g., linear regression) to fit performance. This process allows the scheduler to operate without needing detailed knowledge of each sub-module's implementation.

\textbf{\textit{Back-end}}. The scheduler utilizes the performance models developed for each module. This enables the automatic optimization of task workflows. The back-end, while not delving into detailed programming, recognizes and arranges tasks at the sub-module level. The execution of these tasks is then managed by the \textit{front-end}.

\subsection{Implementation}
We implement our system, FSMoE, atop PyTorch with its
C/C++ and CUDA extension features. For customized algorithms, users can implement their own MoE components by inheriting our abstract interfaces as shown in Listing \ref{list1}. To use the MoE layer, as shown in Listing \ref{list2}, one can instantiate a new instance of an MoE layer, which can be used as a normal nn.Module in PyTorch.

\begin{lstlisting}[caption= Code sample of implementing the abstractions.,label=list1,language=Python]
from FSMoE import ExpertBase, CallbackBase

class CustomizedExpert(ExpertBase):
    def do_experts(self, args):
        pass
        
class CustomizedCallBack(CallbackBase):
    def before_moe_start_hook(self, args):
        pass
\end{lstlisting}

\begin{lstlisting}[language=Python, caption={Code sample of using FSMoE. \texttt{moe\_module} can be used as a normal \texttt{nn.Module} instance in PyTorch.},label=list2,language=Python]
from FSMoE import LinearGate, SimpleOrder, MOELayer

gate_impl = LinearGate()
order_impl = SimpleOrder()
moe_module = MOELayer(gate_impl, order_impl, **kwargs)
\end{lstlisting}

\begin{figure}[!t]
	\centering
	\begin{subfigure}[b]{0.475\textwidth}
		\centering
		\includegraphics[width=\linewidth]{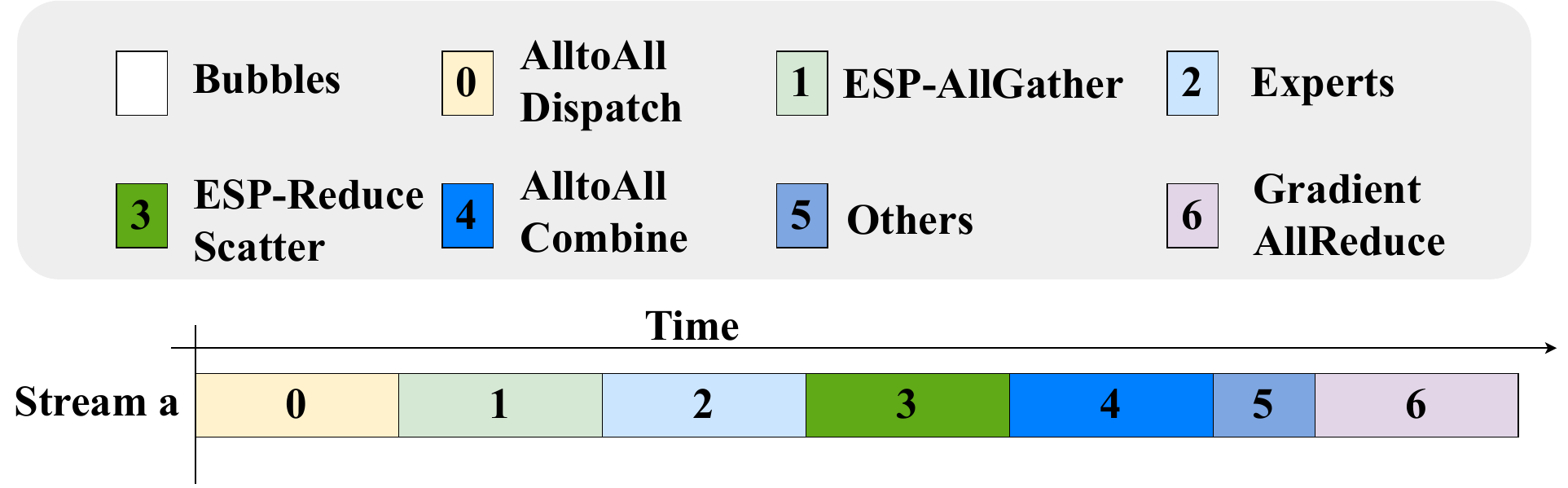}
		\caption{The default schedule (all operations are executed sequentially).}
		\label{fig:nooverlap}
	\end{subfigure}

	\begin{subfigure}[b]{0.475\textwidth}
		\centering
		\includegraphics[width=\linewidth]{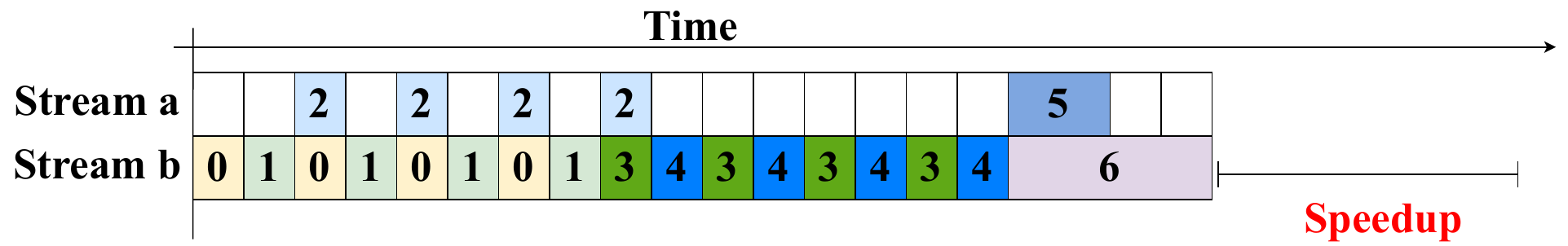}
		\caption{The schedule of Tutel (or PipeMoE) with Gradient-AllReduce overlapped with other dense operations.}
		\label{fig:pipemoe}
	\end{subfigure}
	
	\begin{subfigure}[b]{0.475\textwidth}
		\centering
		\includegraphics[width=\linewidth]{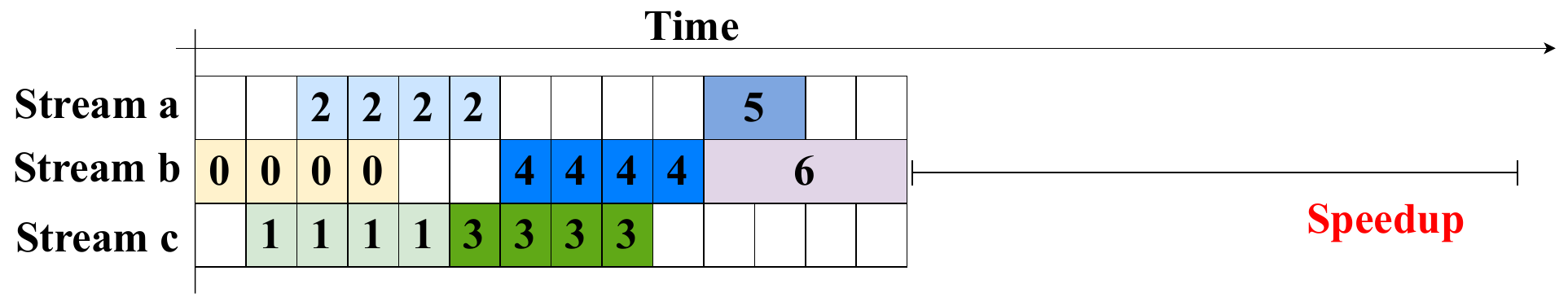}
		\caption{Our proposed schedule FSMoE w/o partitioning the gradient.}
		\label{fig:ourswopg}
	\end{subfigure}
	
	\begin{subfigure}[b]{0.475\textwidth}
		\centering
		\includegraphics[width=\linewidth]{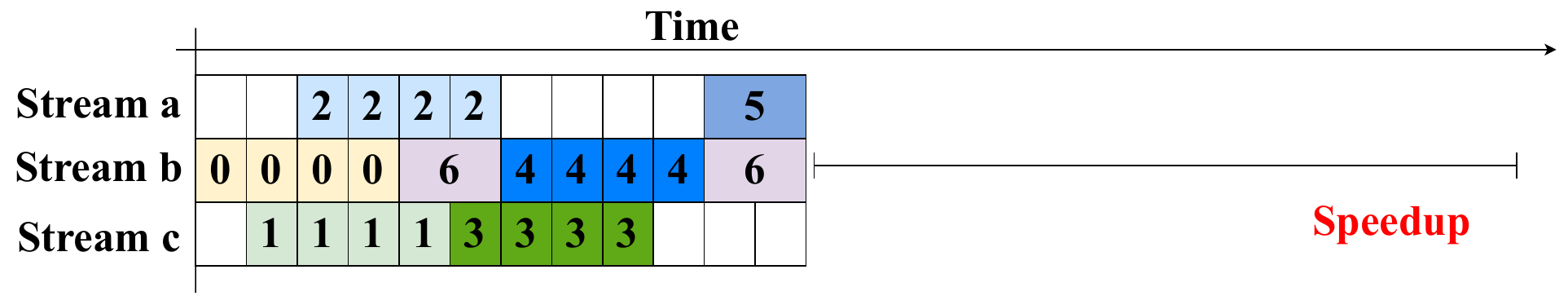}
		\caption{Our proposed schedule FSMoE w/ partitioning the gradient.}
		\label{fig:ours}
	\end{subfigure}
	\caption{Backpropagation of four schedules in DP+MP+EP+ESP with the pipeline degree $r=4$ including (a) the default schedule, (b) an improved Tutel version (Tutel-Improved) where Gradient-AllReduce is overlapped with other dense operations using PipeMoE, (c) our proposed schedule FSMoE without partitioning the gradient, and (d) our proposed schedule FSMoE. The forward process is similar to the backpropagation except for the absence of the Gradient-AllReduce.}
	\label{fig:schedules}
\end{figure}

\section{Optimized Scheduling of Tasks}\label{sec4}
Motivated by the potential overlap between inter-node and intra-node communications, we design a new schedule to pipeline all time-consuming communication tasks (ESP-AllG-
ather, ESP-ReduceScatter, AlltoAll Dispatch/Combine, and Gradient-AllReduce communications) and computing tasks (expert and attention computations) when the group of MP and ESP is aligned with the number of GPUs in a node. In such a scenario, ESP-AllGather and ESP-ReduceScatter are intra-node communications, while AlltoAll Dispatch/Combine and Gradient-Allreduce are inter-node communications. 

\change{This scenario is frequently encountered in practice. With respect to the MoE framework, each layer comprises a limited number of experts, but each expert's model is considerably large, preventing it from fitting entirely on a single GPU. For instance, models like Mixtral-8x7B and Qwen1.5-MoE-A2.7B necessitate dividing an expert across multiple GPUs during training. Meanwhile, considering the training hardware system's topology, inter-node communication (via InfiniBand or Ethernet) generally trails behind the faster intra-node communication methods (such as Shared Memory or NVLink). For instance, contemporary GPU clusters such as Nvidia H100 DGX servers are equipped with eight 200Gb/s network interface cards (NICs), which collectively offer a peak bandwidth of 800Gb/s (equivalent to 100GB/s) for communication between any two nodes. In contrast, the NVLink within a server enables a bandwidth of 900GB/s, illustrating that the bandwidth within a single node is significantly greater than that between nodes. To balance both accuracy and training speed effectively, a practical approach is to align the MP and ESP with the number of GPUs contained within each node. For instance, when training Mixtral-8x7B with settings of $N_{MP} = N_{ESP} = 8$ on servers that feature 8 A100-SXM4-80G GPUs, the approach is exactly feasible. This setup can also be simulated using a simulator~\footnote{https://llm-system-requirements.streamlit.app/}.}


As shown in Fig.~\ref{fig:ours}, the inputs are split into several chunks and sequentially processed in a pipeline. Notably, Gradient-Allreduce is followed by the AlltoAll Dispatch on the last partitioned input as it can also be overlapped with ESP-AllGather/ESP-ReduceScatter and expert computations in the backward phase. The forward phase is similar to the backward phase, except for Gradient-Allreduce. In addition, the optimal pipeline degree varies by phase, necessitating phase-specific solutions. To achieve the new proposed schedule, we first build performance models of different time-consuming computing and communication tasks like PipeMoE~\cite{shi2023pipemoe} and FasterMoE~\cite{he2022fastermoe}. We then formulate an optimization problem based on the performance model and propose an efficient solution.

\begin{figure}[!t]
	\centering
        \begin{subfigure}[t]{0.98\linewidth}
            \centering
            \includegraphics[width=1.0\textwidth]{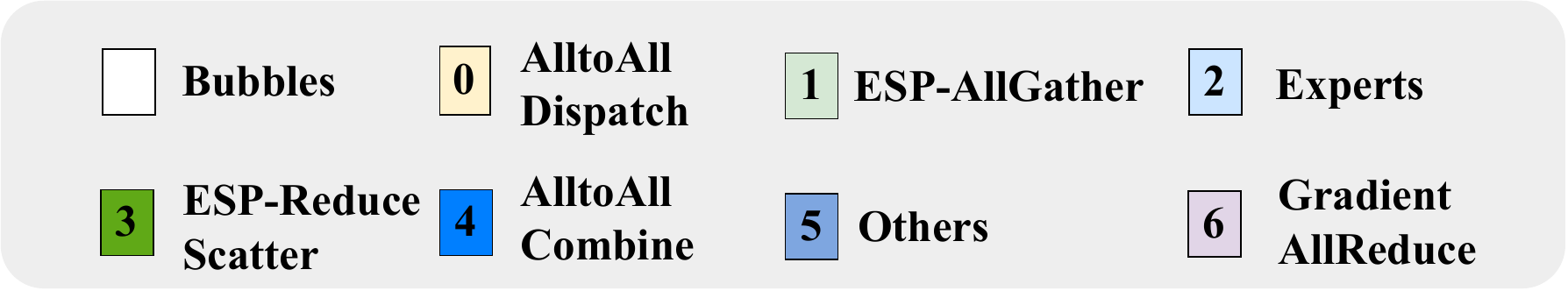}
        \end{subfigure}
        
	\begin{subfigure}[t]{0.48\linewidth}
		\centering
		\includegraphics[width=1.0\textwidth]{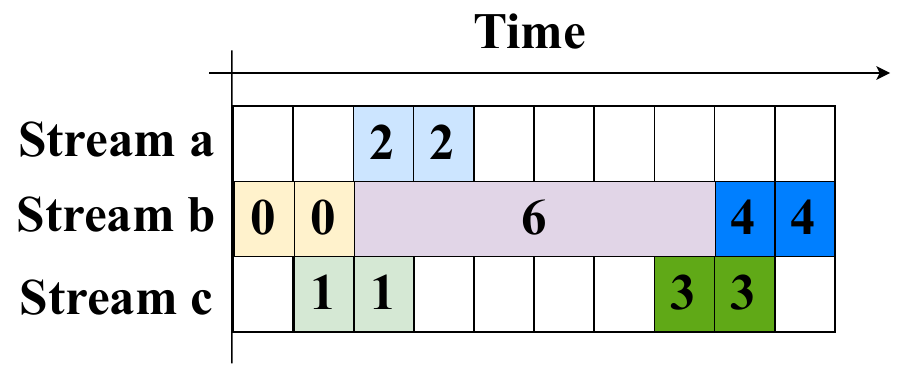}
		\caption{\textbf{Case1.}}
		\label{fig:case1}
	\end{subfigure}
	\begin{subfigure}[t]{0.48\linewidth}
		\centering
		\includegraphics[width=1.0\textwidth]{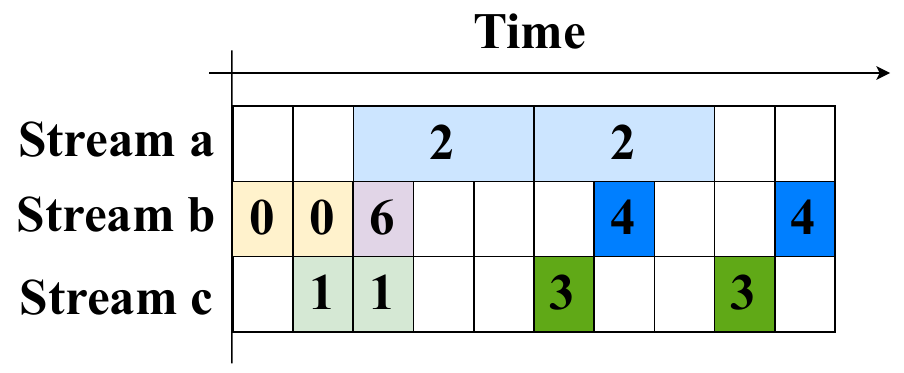}
		\caption{\textbf{Case2.}}
		\label{fig:case2}
	\end{subfigure}
	
	\begin{subfigure}[t]{0.48\linewidth}
		\centering
		\includegraphics[width=1.0\textwidth]{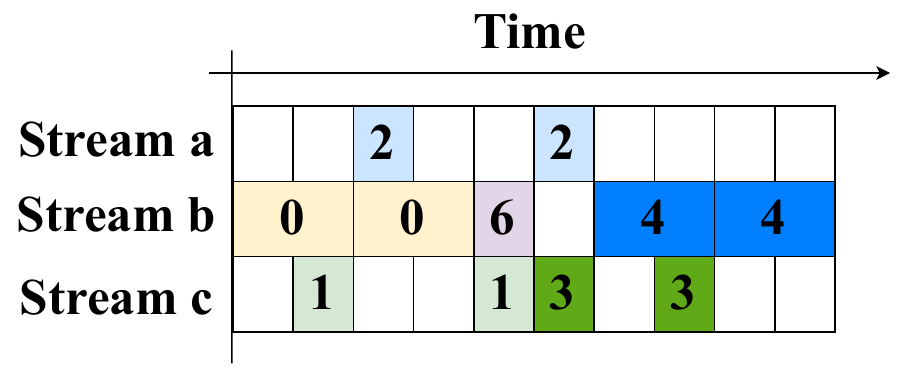}
		\caption{\textbf{Case3.}}
		\label{fig:case3}
	\end{subfigure}
	\begin{subfigure}[t]{0.48\linewidth}
		\centering
		\includegraphics[width=1.0\textwidth]{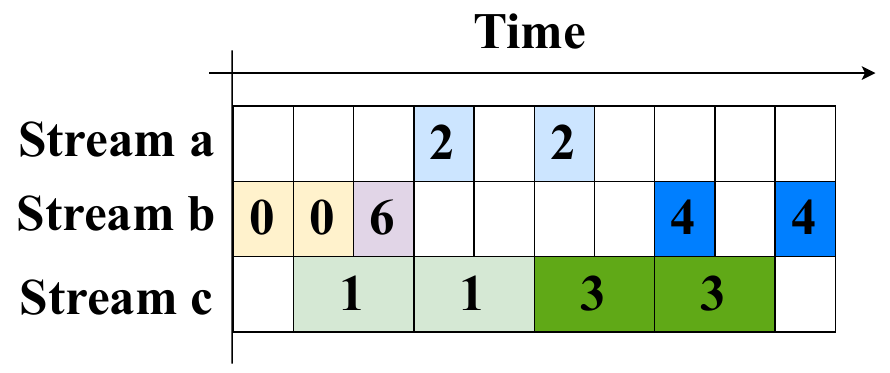}
		\caption{\textbf{Case4.}}
		\label{fig:case4}
	\end{subfigure}
	\caption{Four cases when scheduling the pipelining of ESP-AllGather/ESP-ReduceScatter, AlltoAll Dispatch/Combine, expert computations and Gradient-AllReduce with the pipeline degree $r=2$. \change{(a) \textbf{Case1:} The AlltoAll communications are slower than intra-node communication and expert computations, but the inter-node communications (AlltoAll and Gradient-AllReduce) are not slower than intra-node communication and expert computations.} \change{(b) \textbf{Case2:} Expert computations are not slower than inter-node communications and intra-node communications.} \change{(c) \textbf{Case3:} The AlltoAll communications are not slower than intra-node communication and expert computations. (d) \textbf{Case4:} The intra-node communications (AllGather and ReduceScatter) are not slower than inter-node communications and expert computations.}}
	\label{fig:cases}
\end{figure}

\subsection{Performance Models}\label{sec4.1}
The time required for each chunk in the AlltoAll, AllGather, ReduceScatter, and expert computation processes on inputs divided into $r$ chunks is represented by $t_{a2a, r}$, $t_{ag, r}$, $t_{rs, r}$, and $t_{exp, r}$ respectively. These times are modelled via linear models~\cite{shi2023pipemoe} 
as follows (will verify in \S\ref{subsec:perfmodel-evaluation}):
\begin{equation}\label{pm}
    \begin{array}{l}
{t_{a2a, r}} = {\alpha _{a2a}} + \frac{n_{a2a}}{r} \cdot {\beta _{a2a}},\\
{t_{ag, r}} = {\alpha _{ag}} + \frac{n_{ag}}{r} \cdot {\beta _{ag}},\\
{t_{rs, r}} = {\alpha _{rs}} + \frac{n_{rs}}{r} \cdot {\beta _{rs}},\\
{t_{exp, r}} = {\alpha _{exp }} + \frac{n_{exp}}{r} \cdot {\beta _{exp }},
\end{array}
\end{equation}
where $n_{*}$ represents the volume of the communication message or the computational workload, ${\alpha _{*}}$ denotes the startup time and ${\beta _{*}}$ represents the time per byte transmitted or per unit of workload processed. Particularly, when each expert computation includes multiple identical general matrix-multiplication (GEMM) operations, $\alpha _{exp}$ and $\beta _{exp}$ are determined by multiplying $\alpha _{gemm }$ and $\beta _{gemm}$ by the number of these operations.




\subsection{Optimizing the Pipeline Degree}\label{sec4.2}
The performance model for both computation and communication supports optimizing the pipeline degree $r$ to minimize time costs.

\change{Direct optimization of overall time consumption is challenging because it relies on numerous factors. For instance, the start time of an ESP-ReduceScatter is constrained by both ESP-AllGather (inter-node communication contention) and Expert (data dependence). These constraints complicate finding effective solutions. We classify all general cases into four scenarios, as shown in Fig.~\ref{fig:cases} according to the main source of time consumption in each. For each case, we ease the complexity of the problem by focusing on certain constraints, thereby allowing the optimal solution to be obtained more straightforwardly.} \change{Specifically, (a) \textbf{Case1:} \change{The AlltoAll communications are slower than intra-node communication and expert computations, but the inter-node communications (AlltoAll and Gradient-AllReduce) are not slower than intra-node communication and expert computations.} (b) \textbf{Case2:} \change{Expert computations are not slower than inter-node communications and intra-node communications.} (c) \textbf{Case3:} The AlltoAll communications are not slower than intra-node communication and expert computations. (d) \textbf{Case4:} The intra-node communications (AllGather and ReduceScatter) are not slower than inter-node communications and expert computations.} \change{In situations where multiple time-consuming factors are equally significant, they can be categorized into one of these cases. For instance, when the time consumption for inter-node communication equals that of computation, it can fall into either Case1 or Case2.}
Before discussing these scenarios, the paper formulates seven constraints characterizing these cases, presented as follows.
\begin{center}
\(
\textbf{Q1:} \quad {t_{a2a, r}} > {t_{ag, r}}.
\)
\end{center}
 \textbf{Q1 is True:} implies AlltoAll consumes more time than AllGather for the chunked input. Assuming AllGather and ReduceScatter require similar durations, AlltoAll also exceeds ReduceScatter in time consumption.
\begin{center}
\(
\textbf{Q2:} \quad r \cdot {t_{exp, r}} > 2(r - 1) \cdot {t_{a2a, r}}.
\)
\end{center}
  \textbf{Q2 is True:} indicates that expert computations exceed the duration of communication tasks, excluding AlltoAll Dispatch for the first and AlltoAll Combine for the last chunk. When \textbf{Q1 is True}, this also applies to AllGather and ReduceScatter for the first and last chunks, respectively. 
\begin{center}
\(
\textbf{Q3:} \quad r \cdot {t_{exp, r}} > (r - 1) \cdot \left( {t_{ag, r}} + {t_{rs, r}} \right).
\)
\end{center}
  \textbf{Q3 is True:} means that the time cost of expert computations is large enough to affect the time cost when \textbf{Q1 is False}. 
\begin{center}
\(
\textbf{Q4:} \quad {t_{gar}} > {t_{ag, r}} + {t_{rs, r}}.
\)
\end{center}
  \textbf{Q4 is True:} means that the time cost of Gradient AllReduce is large enough to affect the time cost when \textbf{Q1 is True and Q2 is False}.
\begin{center}
\(
\textbf{Q5:} \quad  {t_{gar}} > r \cdot {t_{exp, r}} - 2(r - 1) \cdot {t_{a2a, r}} + {t_{ag, r}} + {t_{rs, r}}.
\)
\end{center}
  \textbf{Q5 is True:} means that the time cost of Gradient AllReduce is large enough to affect the time cost when \textbf{Q1 is True and Q2 is True}.
\begin{center}
\(
\textbf{Q6:} \quad  {t_{gar}} >  r \cdot {t_{ag, r}} + r \cdot {t_{rs, r}} - 2(r - 1) \cdot {t_{a2a, r}}.
\)
\end{center}
  \textbf{Q6 is True:} means that the time cost of Gradient AllReduce is large enough to affect the time cost when \textbf{Q1 is False and Q3 is False}.
\begin{center}
\(
\textbf{Q7:} \quad  {t_{gar}} > {t_{ag, r}} + {t_{rs, r}} + r \cdot {t_{exp, r}} - 2(r - 1) \cdot {t_{a2a, r}}.
\)
\end{center}
  \textbf{Q7 is True:} means that the time cost of Gradient-AllReduce is large enough to affect the time cost when \textbf{Q1 is False and Q3 is True}.

With these constraints, four cases can be represented as follows:

1) \textbf{Case1: }\textit{(Q1 is True, Q2 is False and Q4 is True) or (Q1 is True, Q2 is True and Q5 is True) or (Q1 is False, Q3 is False and Q6 is True) or (Q1 is False, Q3 is True and Q7 is True)}, which indicates that Gradient-AllReduce is large enough so that the inter-node communications (AlltoAll and Gradient-AllReduce) dominate the time cost in Fig.~\ref{fig:case1}. So we have
\begin{equation}
{t^{moe}_{1}} = 2r \cdot {t_{a2a, r}} + {t_{gar}} = 2r{\alpha_{a2a}} + 2{n_{a2a}} {\beta_{a2a}} + {t_{gar}}.
\end{equation}
Therefore, to find its minima, $t_{1}^{*}$, we should solve
\begin{equation}
\begin{aligned}
    \text{minimize: } & f_1(r) = t^{moe}_1, \\
    \text{s.t. } & r \geq 1, \\
    &(Q1 \land \neg Q2 \land Q4) \lor (Q1 \land Q2 \land Q5) \\
        &\lor (\neg Q1 \land \neg Q3 \land Q6) \lor (\neg Q1 \land Q3 \land Q7).
\end{aligned}
\nonumber
\end{equation}

2) \textbf{Case2: }\textit{(Q1 is True, Q2 is True and Q5 is False) or (Q1 is False, Q3 is True and Q7 is False)}, which indicates that Gradient-Allreduce is too small to influence the time cost and expert computation occupies a dominant position in Fig.~\ref{fig:case2}. So we have
\begin{equation}
\begin{aligned}
{t^{moe}_{2}} =& 2{t_{a2a, r}} + {t_{ag, r}} + {t_{rs, r}} + r \cdot {t_{exp, r}}\\
 =& 2{\alpha_{a2a}} + \frac{2{n_{a2a}}}{r} {\beta_{a2a}} + {\alpha_{ag}} + \frac{{n_{ag}}}{r} {\beta_{ag}} \\
& +{\alpha_{rs}} + \frac{{n_{rs}}}{r}  {\beta_{rs}}  + r{\alpha_{exp}} + {n_{exp}}{\beta_{exp}}.
\end{aligned}
\nonumber
\end{equation}
Therefore, to find its minima, $t_{2}^{*}$, we should solve
\begin{equation}
\begin{aligned}
    \text{minimize: } & f_2(r) = t^{moe}_2, \\
    \text{s.t. } & r \geq 1, \\
    & (Q1 \land Q2 \land \neg Q5) \lor (\neg Q1 \land Q3 \land \neg Q7).
\end{aligned}
\nonumber
\end{equation}
3) \textbf{Case3: }\textit{Q1 is True, Q2 is False and Q4 is False}, which indicates that Gradient-Allreduce and expert computation are too small to influence the time cost. The communications dominate the time cost. And AlltoAll also takes more time than AllGather and ReduceScatter on a chunked tensor in Fig.~\ref{fig:case3}. So we have
\begin{equation}
\begin{aligned}
{t^{moe}_{3}} &= 2r \cdot {t_{a2a, r}} + {t_{ag, r}} + {t_{rs, r}} \\
&= 2r{\alpha_{a2a}} + {2n_{a2a}} {\beta_{a2a}} + {\alpha_{ag}} + \frac{{n_{ag}}}{r} {\beta_{ag}} + {\alpha_{rs}} + \frac{{n_{rs}}}{r} {\beta_{rs}}.
\end{aligned}
\nonumber
\end{equation}
Therefore, to find its minima, $t_{3}^{*}$, we should solve
\begin{equation}
\begin{aligned}
    \text{minimize: } & f_3(r) = t^{moe}_3, \\
    \text{s.t. } & r \geq 1, \\
    & Q1 \land \neg Q2 \land \neg Q4.
\end{aligned}
\nonumber
\end{equation}
4) \textbf{Case4: }\textit{Q1 is False, Q3 is False and Q6 is False}, which indicates that Gradient-Allreduce and expert computation are too small to influence the time cost. And AllGather and ReduceScatter also take more time than AlltoAll on a partitioned tensor. Intra-node communications dominate the time cost in Fig.~\ref{fig:case4}. So we have
\begin{equation}
\begin{aligned}
{t^{moe}_{4}} &= 2{t_{a2a, r}} + r \cdot {t_{ag, r}} + r \cdot {t_{rs, r}}\\
&= 2{\alpha_{a2a}} + \frac{{2n_{a2a}}}{r}  {\beta_{a2a}} + r{\alpha_{ag}} + {n_{ag}}{\beta_{ag}} + r{\alpha_{rs}} + {n_{rs}}{\beta_{rs}}.
\end{aligned}
\nonumber
\end{equation}
Therefore, to find its minima, $t_{4}^{*}$, we should solve
\begin{equation}
\begin{aligned}
    \text{minimize: } & f_4(r) = t^{moe}_4, \\
    \text{s.t. } & r \geq 1, \\
    & \neg Q1 \land \neg Q3 \land \neg Q6.
\end{aligned}
\nonumber
\end{equation}

\subsection{Algorithm}
\begin{algorithm}[!t]
	\caption{FindOptimalPipelineDegree}\label{algo:solve-schedule}
 \small
 \begin{flushleft}
	\textbf{Input: }$\alpha_{a2a}$, $\beta_{a2a}$, $n_{a2a}$,
 $\alpha_{ag}$, $\beta_{ag}$, $n_{ag}$,
 $\alpha_{rs}$, $\beta_{rs}$, $n_{rs}$,
 $\alpha_{exp}$, $\beta_{exp}$, $n_{exp}$, $t_{gar}$ \\
\textbf{Output: }$r$ and $t^{moe}$
\end{flushleft}
	\begin{algorithmic}[1]
        \State $r1,t1=solve(f_1)$ \Comment{Solve with SLSQP}
        \State $r2,t2=solve(f_2)$
        \State $r3,t3=solve(f_3)$
        \State $r4,t4=solve(f_4)$
		\State $\text{candidate\_mins}=[t1,t2,t3,t4]$
        \State $\text{candidates}=[r1,r2,r3,r4]$
        \State $r=\text{candidates}[\text{argmin}(\text{candidate\_mins})]$
        \State $t^{moe}=min(\text{candidate\_mins})$
        \State return $r$ and $t^{moe}$.
	\end{algorithmic}
\end{algorithm}

Algorithm~\ref{algo:solve-schedule} determines the optimal pipeline degree using MoE-related coefficients ($n_{a2a}$, $n_{ag}$, $n_{rs}$, $n_{exp}$) and cluster-related coefficients ($\alpha_{a2a}$,$ \beta_{a2a}$,$ \alpha_{ag}$,$ \beta_{ag}$,$ \alpha_{rs}$,$ \beta_{rs}$,$ \alpha_{exp}$,$ \beta_{exp}$). In particular, $t_{gar}$ is a manually entered value that is set to zero in the forward process and determined by \textsection\ref{sec5} in the backward process. FSMoE supports varied pipeline degrees in both phases. The algorithm executes once before training, following the estimation of cluster-related coefficients. The ``solve'' function employs a sequential least squares programming (SLSQP)~\cite{nocedal1999numerical} solver. This algorithm is quadratic convergence in solving $f_1$, $f_2$,
$f_3$ and $f_4$ (Lines 1-4), and other operations take O(1) time
complexity.

\subsection{Schedule Forward and Backward Separately}\label{sec4.4}
Because of the calculation of gradient w.r.t. the weight and the gradient synchronization among DP workers, the tasks in backpropagation are different from the forward phrase. The optimal pipeline degree thus differs. Therefore, we manually implement the backpropagation by storing the activation of each computational operation and computing the gradient. 

Specifically, the parameters $\alpha_{exp}$, $\beta_{exp}$, and $n_{exp}$ in the backward phase are twice those in the forward phase to accommodate the derivatives of both weight and input. Meanwhile, $t_{gar}$ is set to zero in the forward phase as gradient synchronization does not occur, and it is determined by the algorithm detailed in \textsection\ref{sec5} for the backward phase.

\section{Scheduling for Backpropagation}\label{sec5}
Due to the inter-node communication in the MoE layer, Gradient-AllReduce of the gradient synchronization can not be directly overlapped with MoE layers. A dedicated co-design is necessary to further hide the time cost of Gradient-AllReduce. We propose to adaptively partition the gradients to achieve the maximal overlap of Gradient-AllReduce with other operations.

Our approach contains two steps. Step 1: We calculate the time cost of the parts that can be overlapped with Gradient-AllReduce (denoted as \textit{overlappable parts}) for all layers. Then we slice the gradient and assign them to these overlap-able parts as far as possible. Step 2: We arrange the remaining gradient after the first step and set the remaining gradient partitioned to each MoE layer as variables to optimize the assignment.

\subsection{Performance Model}
Similar to \textsection\ref{sec4.1}, the performance model of AllReduce can be represented as $ {t_{ar}}({n_{ar}}) = {\alpha _{ar}} + {n_{ar}} \cdot {\beta _{ar}} $,
where $t_{ar}$ denotes the elapsed-time, $n_{ar}$ represents the amount of the communication message, ${\alpha _{ar}}$ denotes the starup time and  ${\beta _{ar}}$ represents the transmission time per byte. The inverse function of $ {t_{ar}}({n_{ar}})$ is represented as ${g_{grad}^{inv}}(t_{ar}) = (t_{ar}-{\alpha _{ar}})/ {\beta _{ar}}$.

\subsection{Step 1: Calculate Partitioned Gradients}
In this step, we first optimize the pipeline degree of each MoE layer with $t_{gar}=0$ by Algorithm \ref{algo:solve-schedule} to calculate the time cost of \textit{overlappable parts}. Then we try to slice the gradient and assign them to \textit{overlappable parts} of each layer in order to minimize the training time. According to the above performance model, we are able to calculate the gradient assigned to each layer. 

For convenience, we denote an MoE layer and other operations before the next MoE layer as a generalized layer. We denote the gradient for a generalized layer $i$ as $n_{grad}^i$ and the time cost of \textit{overlappable parts} as $t_{olp}^i$. The number of gradients assigned to each layer in this step can be represented as
\begin{equation}
    n_{first}^i = g_{grad}^{inv}(\min ({t_{grad}}(n_{grad}^{i - 1}),t_{olp}^i)).
\end{equation}
If $n_{grad}^{i - 1}$ is not fully overlapped, $n_{grad}^i$ should be updated by
\begin{equation}
   n_{grad}^i = n_{grad}^i + g_{grad}^{inv}(\max ({t_{grad}}(n_{grad}^{i - 1}) - t_{olp}^i,0)).
\end{equation}
Notably, the time cost of \textit{overlappable parts}, $t_{olp}$, can be divided into sparse MoE parts $t_{olp,moe}$ and dense parts $t_{olp,dense}$. The dense parts $t_{olp,dense}$ can be measured before the training, while $t_{olp,moe}$ can be calculated during the optimization of the pipeline degree. Specifically, when $t_{gar}=0$, we will encounter Case2, Case3 and Case4 mentioned in \textsection\ref{sec4.2}. And $t_{olp,moe}$ can be formulated as following
\begin{equation}
t_{olp,moe}(r) = \begin{cases}
r \cdot t_{exp, r} + t_{ag, r} + t_{rs, r}  - 2(r - 1)t_{a2a, r}, & \text{Case 2} \\
t_{ag, r} + t_{rs, r}, & \text{Case 3} \\
r \cdot t_{ag, r} + r \cdot t_{rs, r}  - 2(r - 1)t_{a2a, r}, & \text{Case 4}
\end{cases}
\nonumber
\end{equation}
After the above process, we will enter the second step if gradients still remain.

\subsection{Step 2: Optimize Partitioning}
The second step is to assign the remaining gradients after the first step. Note that with different input time costs of Gradient-AllReduce, the optimization algorithm (Algorithm~\ref{algo:solve-schedule}) would produce different degrees and time costs. It indicates that the remaining gradients can be further partitioned into MoE layers to minimize the training time. 

We denote the remained gradient for the generalized layer $i$ as $n_{rem}^i$ and the Algorithm \ref{algo:solve-schedule} as $f_{moe}^i(t_{gar})$ who takes the time cost of Gradient-AllReduce as the input and produces the time cost of the MoE layer $i$. Then, set the remaining gradient assigned to the MoE layer $i$ as $x_g^i$. The optimization model can be represented as
\begin{equation}
    \text{minimize: }{f_g}({X_g}) = \sum\limits_{i = 1}^{{n_l}} {f_{moe}^i\left( {{t_{grad}}(x_g^i)} \right)},
    \nonumber
\end{equation}
\begin{equation}
    \text{s.t. }0 \le x_g^i < n_{rem}^i + \sum\limits_{j = i - 1}^{{n_l}} {(n_{rem}^j - x_{gar}^j)} ,0 < i < {n_l},
\end{equation}
where ${n_l}$ represents the number of layers. As the optimization will be conducted only once before the training, we do not need to care too much about the time cost. Therefore, we simply adopt the differential evolution algorithm~\cite{price2013differential} when we solve the above optimization problem.

\begin{table}[]
	\centering
 \addtolength{\tabcolsep}{-3pt}
 		\caption{The server configurations in our testbeds.}
		\label{tab:server-config}
   \resizebox{\linewidth}{!}{%
\begin{tabular}{lll}
\hline
\multicolumn{1}{|l|}{Name}    & \multicolumn{1}{l|}{Testbed A}                                                             & \multicolumn{1}{l|}{Testbed B}                                                          \\ \hline
 \hline
\multicolumn{1}{|l|}{CPU}     & \multicolumn{1}{l|}{Dual Intel(R) Xeon(R) Platinum}                                                 & \multicolumn{1}{l|}{Dual Intel(R) Xeon(R) Gold}                                              \\
\multicolumn{1}{|l|}{}        & \multicolumn{1}{l|}{\begin{tabular}[c]{@{}l@{}} 8358 CPU @ 2.60GHz  \end{tabular}} & \multicolumn{1}{l|}{\begin{tabular}[c]{@{}l@{}} 6230 CPU  @ 2.10GHz\end{tabular}} \\ \hline
\multicolumn{1}{|l|}{GPU}     & \multicolumn{1}{l|}{8x Nvidia RTXA6000 @1.46GHz}                                                    & \multicolumn{1}{l|}{4x Nvidia RTX2080Ti @1.35GHz}                                                \\
\multicolumn{1}{|l|}{}        & \multicolumn{1}{l|}{\begin{tabular}[c]{@{}l@{}} 48GB Mem\end{tabular}}           & \multicolumn{1}{l|}{\begin{tabular}[c]{@{}l@{}} 11GB Memory\end{tabular}}        \\ \hline
\multicolumn{1}{|l|}{Memory}  & \multicolumn{1}{l|}{512GB DDR4}                                                            & \multicolumn{1}{l|}{512GB DDR4}                                                         \\ \hline
\multicolumn{1}{|l|}{NVlink}  & \multicolumn{1}{l|}{112.5GB/s (4x)}                                                             & \multicolumn{1}{l|}{-}                                                                  \\ \hline
\multicolumn{1}{|l|}{PCIe}    & \multicolumn{1}{l|}{4.0 (x16)}                                                             & \multicolumn{1}{l|}{3.0 (x16)}                                                          \\ \hline
\multicolumn{1}{|l|}{Network} & \multicolumn{1}{l|}{Mellanox MT28908 @ 200Gb/s}                                                                     & \multicolumn{1}{l|}{Mellanox MT27800 @ 100Gb/s}                                           \\ \hline
\end{tabular}%
}
\end{table}

\begin{table}[t]
	\centering
	\caption{Configurations of attention and MoE layers. $N_{\text{hscale}}=H/M$. $f=*$ means tokens will not be dropped when gating. \textit{ffn-type} means the type of experts in MoE.}
	\label{tab:moe-configs}
\begin{tabular}{cc}
\hline
\multicolumn{1}{|c|}{-}                   & \multicolumn{1}{c|}{Candidate Values}     \\ \hline
 \hline
\multicolumn{1}{|c|}{$B$}                 & \multicolumn{1}{c|}{\{1,2,4\}}            \\ \hline
\multicolumn{1}{|c|}{$N_{\text{heads}}$}  & \multicolumn{1}{c|}{\{8,16,32\}}          \\ \hline
\multicolumn{1}{|c|}{$L$}                 & \multicolumn{1}{c|}{\{512,1024,2048\}/\{256,512,1024\}}     \\ \hline
\multicolumn{1}{|c|}{$M$}                 & \multicolumn{1}{c|}{\{1024, 2048, 4096\}} \\ \hline
\multicolumn{1}{|c|}{$N_{\text{hscale}}$} & \multicolumn{1}{c|}{\{2,3,4\}}            \\ \hline
\multicolumn{1}{|c|}{$f$}                 & \multicolumn{1}{c|}{\{1.2,2.4,*\}} \\ \hline
\multicolumn{1}{|c|}{\text{ffn-type}}          & \multicolumn{1}{c|}{\{simply,Mixtral\}}    \\ \hline
\end{tabular}
\end{table}

\section{EVALUATION}\label{sec:evaluation}
\subsection{Experimental Settings}

We mainly compare our FSMoE with Tutel~\cite{hwang2023tutel} (w/ its optimized version PipeMoE~\cite{shi2023pipemoe}) which designs an adaptive schedule to determine the pipeline degree of the overlaps, with a focus on pipelining communications and computations in a typical structure of the MoE model in DP+MP+EP+
ESP shown in Fig.~\ref{fig:full_example}. Additionally, we compare the end-to-end training performance of FSMoE with DeepSpeed-MoE~\cite{aminabadi2022deepspeed,DBLP:Rajbhandari22deepspeedmoe}, which is a dedicated MoE training system. The code we implemented is accessible at https://github.com/xpan413/FSMoE.

\textbf{Testbeds}: Experiments are carried out on two distinct testbeds: Testbed-A, a 48-GPU cluster comprising six interconnected nodes, and each node is equipped with four Nvidia A6000 GPUs. Testbed-B, a 32-GPU cluster comprising eight interconnected nodes, and each node is equipped with four Nvidia GeForce RTX2080Ti GPUs. 
More details on the server configuration can be found in Table~\ref{tab:server-config}. The software environments are Ubuntu-20.04, CUDA-11.3, PyTorch-1.12 and NCCL-2.12.

\textbf{MoE model configurations.} We select a combination of input parameters whose ranges are shown in Table~\ref{tab:moe-configs} to cover a variety of typical configurations of attention and MoE layers. $L$ is set to \{256, 512, 1024\} on Testbed-B due to the memory limit of 2080Ti. Notably, we select a range of $N_{\text{hscale}}=H/M$ rather than directly setting $H$, which is more common in real-world scenarios. $f=*$ means tokens will not be dropped when gating. \textit{ffn-type} means the type of experts in MoE. \textit{simple} represents the conventional two feed-forward dense layers and \textit{Mixtral} means the experts using in Mixtral~\cite{jiang2024mixtral}. Additionally, $N_{MP}$ and $N_{ESP}$ are both set to 4 in Testbed-B where ESP-AllGather and ESP-ReduceScatter are intra-node communications while AlltoAll and Gradient-AllReduce are inter-node communications. Similarly, $N_{MP}$ and $N_{ESP}$ are both set to 8 in Testbed-A. 

\subsection{Performance Model}\label{subsec:perfmodel-evaluation}
\begin{figure}[!t]
	\centering
	\begin{minipage}[b]{0.22\textwidth}
		\centering
		\includegraphics[width=\textwidth]{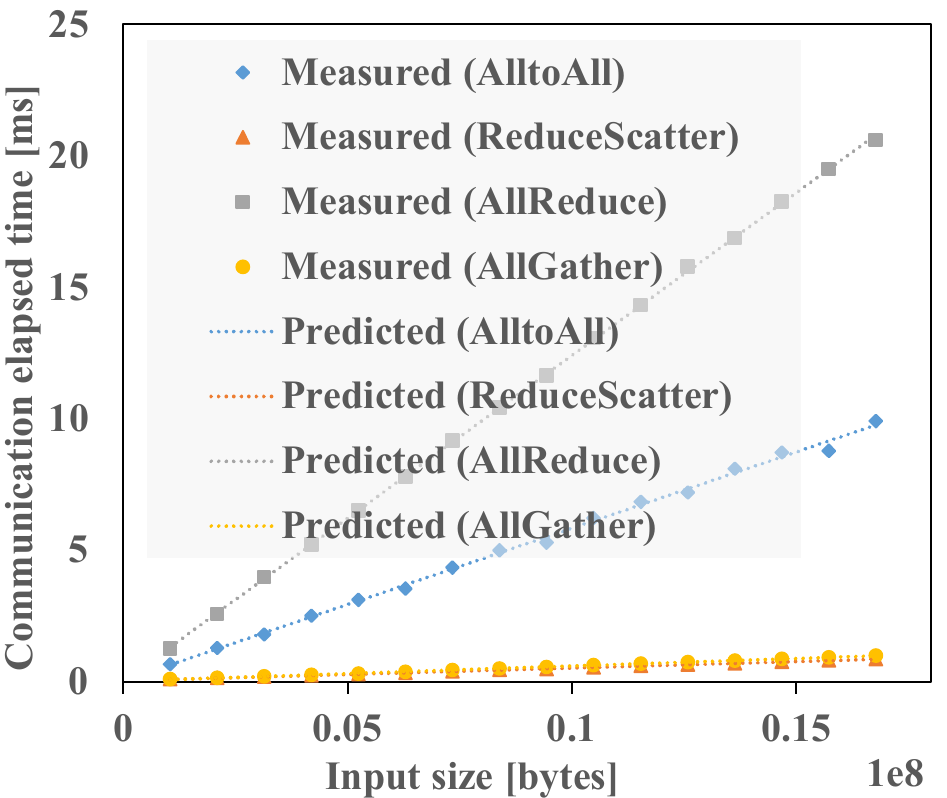}
		\caption*{(a) Communication (A6000).}
		
	\end{minipage}
	\hfill
	\begin{minipage}[b]{0.23\textwidth}
		\centering
		\includegraphics[width=\textwidth]{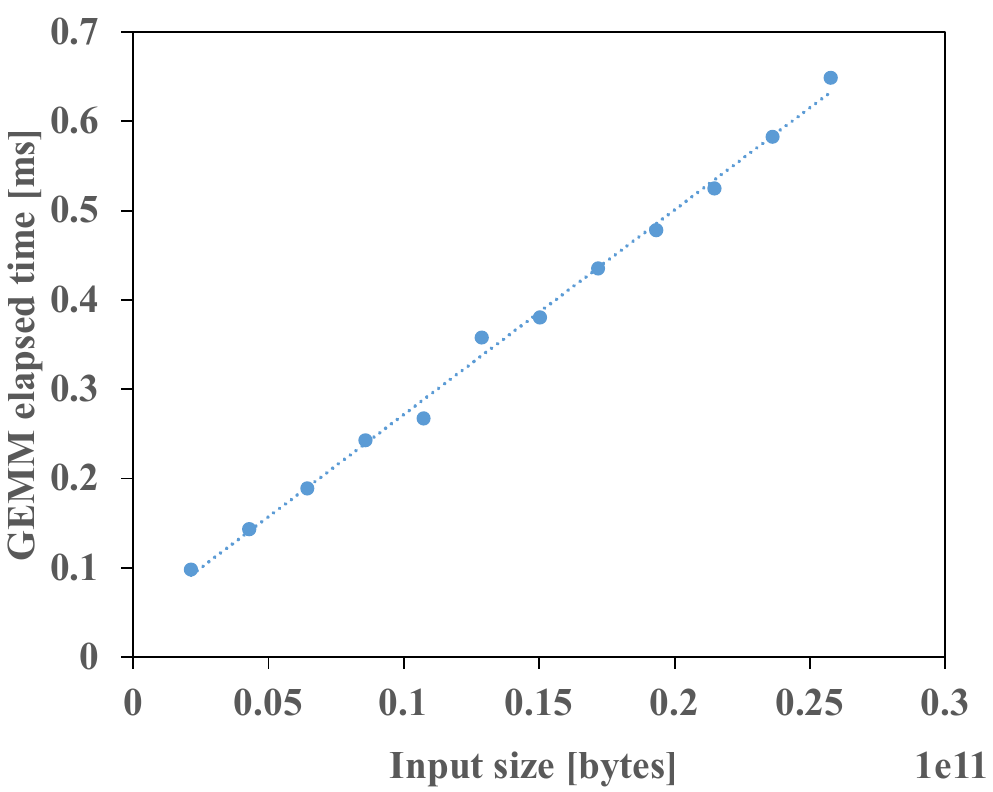}
		\caption*{(b) GEMM (A6000).}
	
	\end{minipage}
	
	\vspace{12pt}
	
	\begin{minipage}[b]{0.23\textwidth}
		\centering
		\includegraphics[width=\textwidth]{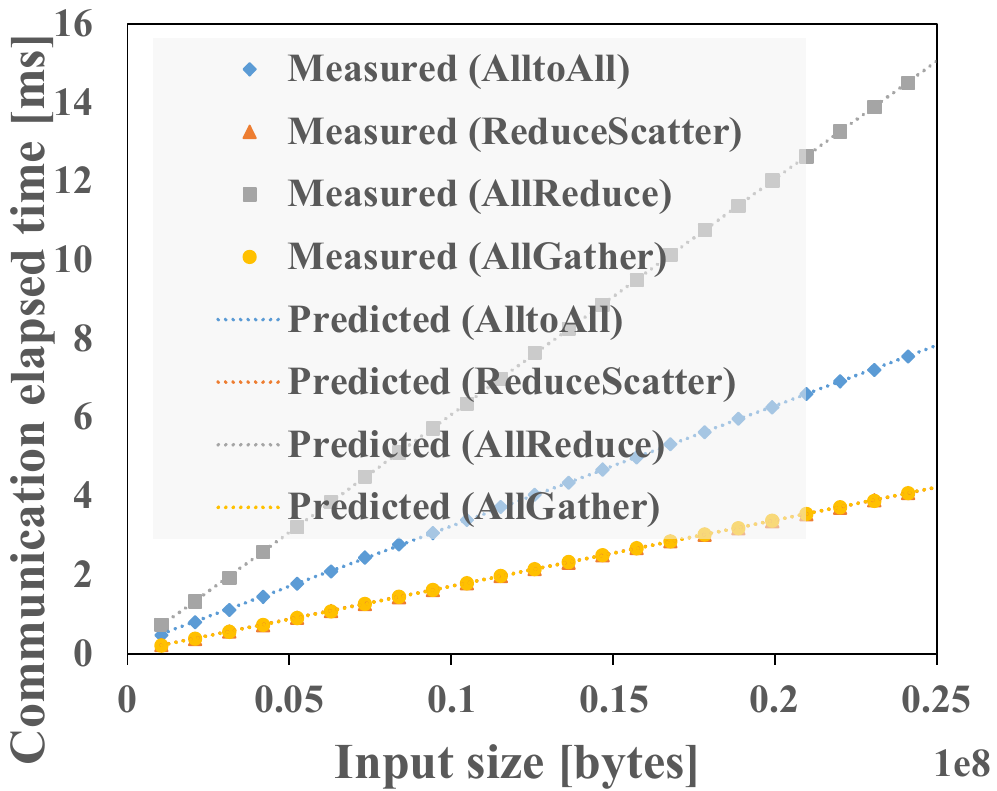}
		\caption*{(c) Communication (2080Ti).}
	
	\end{minipage}
	\hfill
	\begin{minipage}[b]{0.23\textwidth}
		\centering
		\includegraphics[width=\textwidth]{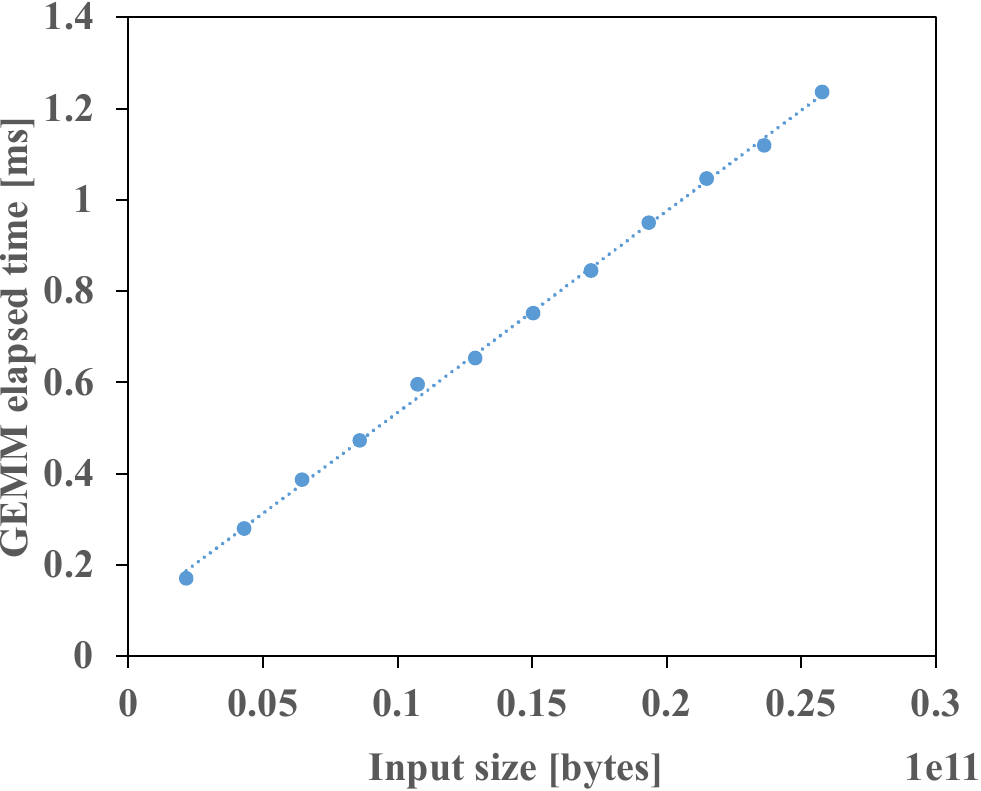}
		\caption*{(d) GEMM (2080Ti).}

	\end{minipage}
	
	\caption{Performance models. Markers are measured values and lines are predicted values with estimated parameters. (a) $\alpha_{gemm}$=\text{4.26e-2}, $\beta_{gemm}$=\text{2.29e-11} on Testbed-A. (b) $\alpha_{a2a}$=\text{2.87e-1}, $\beta_{a2a}$=\text{2.21e-7}, $\alpha_{ag}$=\text{3.37e-1}, $\beta_{ag}$=\text{2.32e-06}, $\alpha_{rs}$=\text{3.95e-1}, $\beta_{rs}$=\text{2.34e-7}, $\alpha_{ar}$=\text{5.11e-1}, $\beta_{ar}$=\text{4.95e-6} on Testbed-A. (c) $\alpha_{gemm}$=\text{9.24e-2}, $\beta_{gemm}$=\text{4.42e-11} on Testbed-B. (d) $\alpha_{a2a}$=\text{1.75e-1}, $\beta_{a2a}$=\text{3.06e-7}, $\alpha_{ag}$=\text{3.20e-2}, $\beta_{ag}$=\text{1.68e-7}, $\alpha_{rs}$=\text{3.91e-2}, $\beta_{rs}$=\text{1.67e-7}, $\alpha_{ar}$=\text{8.37e-2}, $\beta_{ar}$=\text{5.99e-7} on Testbed-B.}
	\label{fig:alphbeta}
 \vspace{-5mm}
\end{figure}

We require the input parameters that are related to the cluster for the performance models of computation and communication. \change{We measure the elapsed time with a range of sizes for GEMM computation and four types of communication to fit the performance models in Eq.~\ref{pm} using microbenchmark tools. In particular, we utilize the NCCL-2.12 collective communication primitives along with \textit{nccl-tests}\footnote{\url{https://github.com/NVIDIA/nccl-tests}} to evaluate communication durations across diverse message sizes. Meanwhile, we employ the \textit{torch.matmul}\footnote{\url{https://pytorch.org/docs/stable/generated/torch.matmul.html}} function in PyTorch to assess the GEMM execution times for matrices of varying shapes. For communication modeling, float-type elements are chosen in a range from $2^{18}$ to $24 \times 2^{18}$, with steps of $2^{18}$, to simulate different tensor sizes. Likewise, for the GEMM modeling, float-type elements are picked from a range between $2^{19}$ and $12 \times 2^{19}$, with $2^{19}$ increments. Each measurement is averaged over five runs to ensure consistency. The results are shown in Fig.~\ref{fig:alphbeta}. It is seen that
our linear models with intercept terms (i.e., startup time) can
well fit the measured performance. Specifically, the \(r^2\) for our GEMM model is 0.9987, and the corresponding \(r^2\) for the communication tasks are as follows: AllReduce: 0.9999896, AlltoAll: 0.9999, AllGather: 0.9999653, and ReduceScatter: 0.9999599. The total time required for both computation and communication in the performance models is under 100 seconds. Fitting through the least squares method takes under 10ms. Following fitting, the empirical time cost for SLSQP in solving \(r\) averages 193ms over 1458 configured cases.
When dealing with a new GPU cluster, it is only necessary to estimate the parameters one time using micro-benchmarks prior to model training, without impacting the training efficiency. }

\begin{table}[htb]
	\centering
	\caption{Averaged speedups of four schedules over Tutel (w/ its optimized version PipeMoE) on configured layers in Table~\ref{tab:moe-configs}. Tutel-Improved means using PipeMoE with Gradient-AllReduce overlapped with non-MoE parts, while FSMoE-No-IIO indicates using FSMoE without the overlaps between inter and intra node communications.}
	\label{tab:all-moelayer-experiment}
\begin{tabular}{lcc}
\hline
\multicolumn{1}{|c|}{\multirow{2}{*}{Schedule}} & \multicolumn{2}{c|}{Speedup}                                          \\ \cline{2-3} 
\multicolumn{1}{|c|}{}                          & \multicolumn{1}{c|}{Testbed-A}     & \multicolumn{1}{c|}{Testbed-B}     \\ \hline
 \hline
\multicolumn{1}{|l|}{Tutel}                   & \multicolumn{1}{c|}{$1.00\times$} & \multicolumn{1}{c|}{$1.00\times$} \\ \hline
\multicolumn{1}{|l|}{Tutel-Improved}             & \multicolumn{1}{c|}{$1.09\times$} & \multicolumn{1}{c|}{$1.08\times$} \\ \hline
\multicolumn{1}{|l|}{FSMoE-No-IIO}                & \multicolumn{1}{c|}{$1.12\times$} & \multicolumn{1}{c|}{$1.16\times$} \\ \hline
\multicolumn{1}{|l|}{FSMoE}                      & \multicolumn{1}{c|}{$1.18\times$} & \multicolumn{1}{c|}{$1.22\times$} \\ \hline
\end{tabular}
\end{table}

\subsection{Performance on Configured Layers}
We conducted a comparison between our proposed method FSMoE and PipeMoE~\cite{shi2023pipemoe} in the structure illustrated in Fig.~\ref{fig:full_example}, using various configurations as outlined in Table~\ref{tab:all-moelayer-experiment}. Notably, the gradient aggregation of a configured layer is added in order to validate our gradient partitioning method and schedule to overlap Gradient-AllReduce.
For better comparison, experiments on two additional schedules (Tutel-Improved and FSMoE-No-IIO) are further conducted. Tutel-Improved means PipeMoE with Gradient-AllReduce overlapped with non-MoE parts, while FSMoE-No-IIO means FSMoE without the overlaps between inter-node and intra-node communications. The experimental results indicate that with a simple overlap between Gradient-AllReduce with non-MoE parts, we can achieve a speed up of $1.08\times$ to $1.09\times$ over Tutel (w/ PipeMoE). And with our gradient partitioning and well overlaps among inter-node and intra-node communication as well as computation tasks, FSMoE achieves an average speedup of $1.18\times$ to $1.22\times$ over Tutel across 1458 cases. By comparing the speed up of our FSMoE and FSMoE-No-IIO over Tutel in Table~\ref{tab:all-moelayer-experiment}, we see that the overlaps between inter-node and intra-node communications further improve the performance.

		
	
	


\begin{figure}[!t]
	\centering
\includegraphics[width=1.0\linewidth]{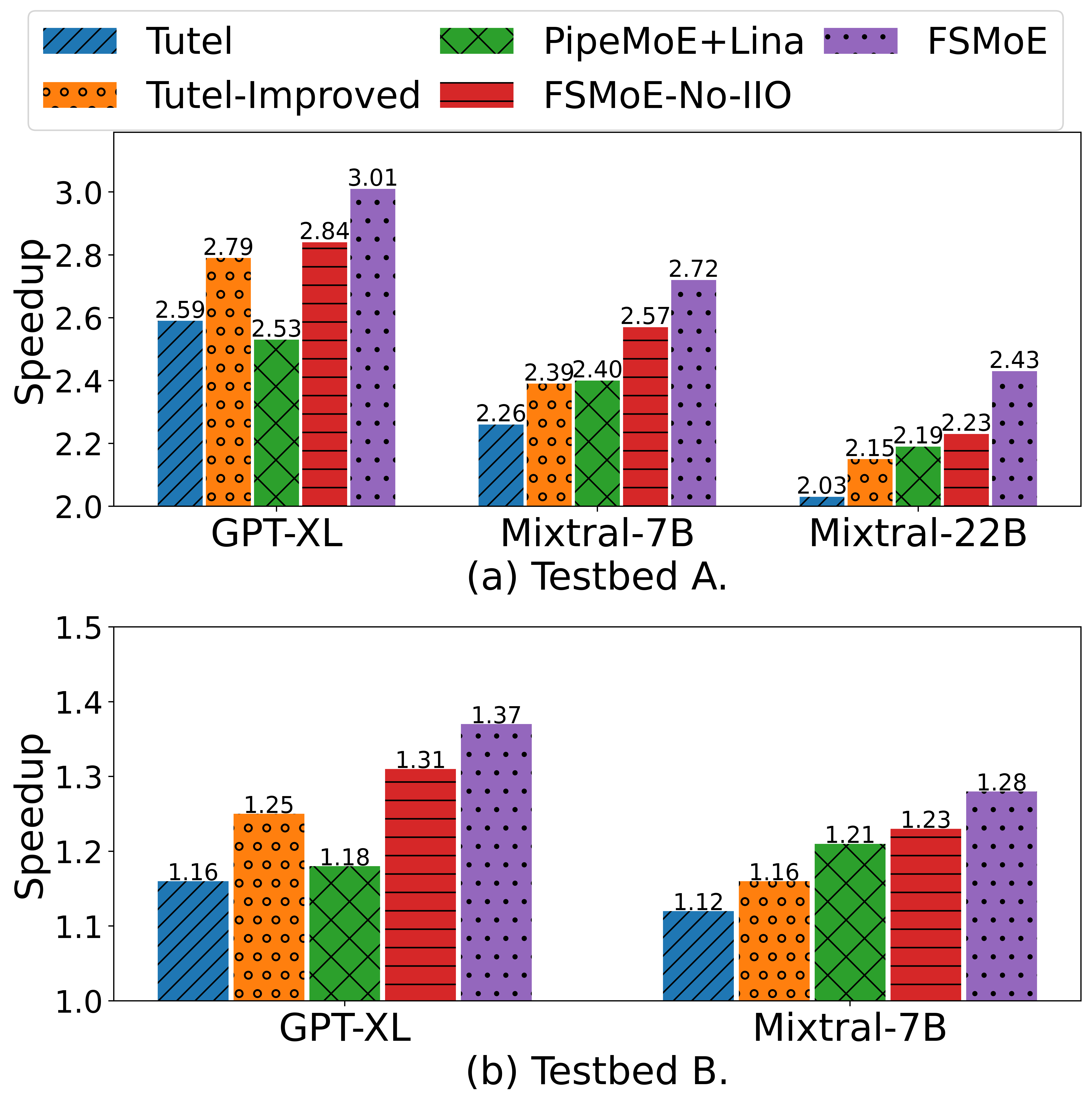}	\caption{Speedups of FSMoE, FSMoE-No-IIO, Tutel, Tutel-Improved, PipeMoE+Lina (PipeMoE with the additional schedule introduced by Lina~\cite{li2023lina} that partitions the gradient into fixed chunk size) over DeepSpeed-MoE (DS-MoE) on MoE models (GPT2-XL, Mixtral-7B and Mixtral-22B). }
 \label{fig:realworld}
\end{figure}

\begin{figure}[!t]
	\centering
\includegraphics[width=1.0\linewidth]{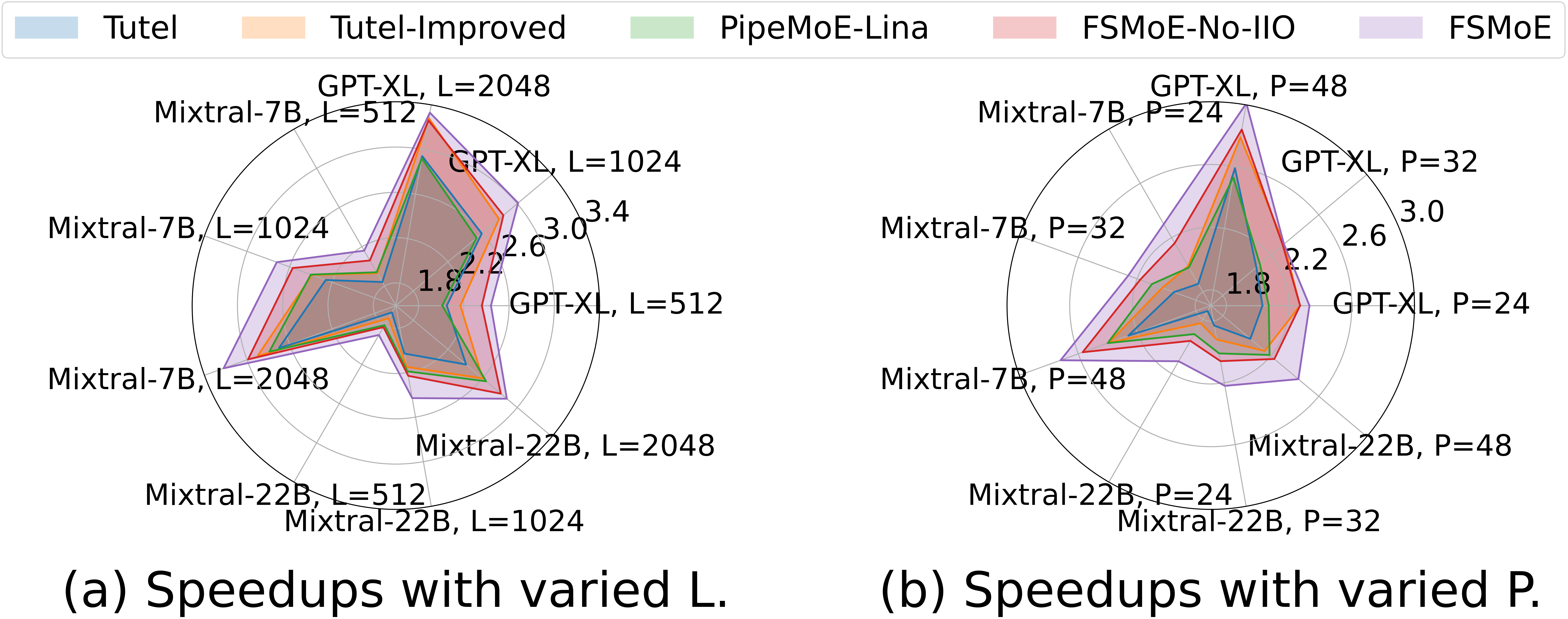}
	\caption{Speedups of five schedules over DS-MoE on Testbed-A with different configurations.}
 \label{fig:realworld_configure}
\end{figure}

\begin{figure}[!t]
	\centering
\includegraphics[width=1.0\linewidth]{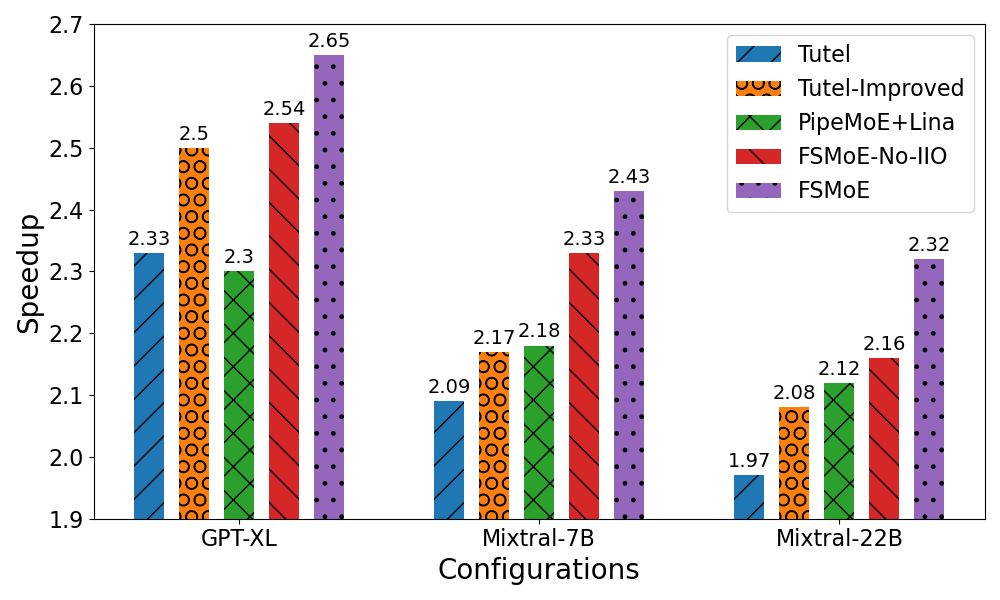}
	\caption{Speedups of five schedules over DS-MoE on Testbed-A when PP is enabled.}
 \label{fig:realworld_pp}
\end{figure}
\subsection{Performance on MoE Models}
To evaluate the end-to-end training performance, we conduct experiments with Mixtral-7B~\cite{jiang2024mixtral} and an MoE model based on GPT-2~\cite{radford2019language} on two testbeds. In addition, experiments in Mixtral-22B are also conducted on Testbed-A. We set $B=1$, $k=2$, $f=1.2$ during the experiment. To enable the overlap between inter and intra communication, $N_{ESP}=N_{MP}$, which is 8 and 4 on Testbed-A and Testbed-B, respectively. Furthermore, the number of experts ($N_{EP}$) is the same as the number of nodes, which is 6 and 8 on Testbed-A and Testbed-B, respectively. $L$ is set to 256 on Testbed-B and to 1024 on Testbed-A. Ensuring the models to be held on Testbed-B (32x 2080Ti 11GB), we set the number of layers for Mixtral-7B to 7. Due to the memory limit, the number of layers for Mixtral-22B is set to 33 on Testbed-A. 
For further analysis, experiments on two additional schedules are conducted. Tutel-Improved indicates Tutel with the overlaps between Gradient-AllReduce with non-MoE parts using PipeMoE. PipeMoE+Lina means PipeMoE with the additional schedule introduced by Lina~\cite{li2023lina} that partitions the gradient into fixed chunk size (e.g., 30MB) and overlaps the partitioned gradient aggregation with expert computations and non-MoE parts in backpropagation. 

The results in Fig.~\ref{fig:realworld} indicate that FSMoE achieves a speedup of $1.28\times$ to $3.01\times$ compared to DeepSpeed-MoE (DS-MoE) while Tutel can only achieve a speedup of $1.16\times$ to $2.59\times$. Additionally, FSMoE can achieve an average speedup of $1.19\times$ over Tutel, $1.12\times$ over Tutel-Improved, $1.14\times$ over PipeMoE+Lina and $1.07\times$ over FSMoE-No-IIO, which validates the efficiency of our adaptive gradient partitioning method and pipelining schedule. It is worth mentioning that Lina's idea of partitioning gradients and scheduling the gradient aggregation can not handle various configurations due to the fixed chunk size. Thus, its performance is hit or miss. And our FSMoE can adaptively partition the gradient and adjust the pipelining degree to achieve better results. 

\textbf{Performance on MoE Models With PP Enabled.}
We also conduct experiments on Testbed-A when PP is further enabled ($N_{PP}=2$), implemented using GPipe~\cite{DBLP:huang2019gpipe}. The results are shown in Fig~\ref{fig:realworld_pp}. The results indicate that FSMoE can achieve an average speedup of $2.46\times$ over DS-MoE, $1.16\times$ over Tutel, $1.10\times$ over Tutel-Improved, $1.12\times$ over PipeMoE+Lina and $1.05\times$ over FSMoE-No-IIO. 

\textbf{Performance on MoE Models with Varied $L$ and Varied $P$.} 
Moreover, we analyze the performance of FSMoE with varied $L$ and $P$ on Testbed-A. $L$ is varied in \{512, 1024 and 2048 \} while $P$ is varied in \{16, 32 and 48 \}. The results are shown in Fig~\ref{fig:realworld_configure}. The results indicate that FSMoE can achieve an average speedup of $2.17\times, 2.72\times \text{ and } 3.14\times$ over DS-MoE
and $1.17\times, 1.19\times \text{ and } 1.17\times$ over Tutel when $L$ is varied in \{512, 1024 and 2048 \} and $P=48$. FSMoE can achieve an average speedup of $2.25\times, 2.27\times \text{ and } 2.72\times$ over DS-MoE, $1.20\times, 1.16\times \text{ and } 1.19\times$ over Tutel when $P$ is varied in \{16, 32 and 48 \} and $L=1024$. It indicates the robustness of FSMoE.



\textbf{Support Multiple Gating Functions.} Table~\ref{tab:more_gate} underscores the ability of our framework to support multiple gating functions while maintaining improved efficiency. Our framework shows potential scalability and flexibility in handling complex MoE architectures.

\begin{table}[!ht]
\centering
	\caption{Time performance on Testbed-B (average iteration time in milliseconds) of various gating on real-world MoE GPT2-XL. The lower is better. Speedup are provided in parentheses.}
\begin{tabular}{|c|cc|}
\hline 
Gating & \multicolumn{1}{c|}{DeepSpeed-MoE} & FSMoE \\ \hline \hline
Gshard~\cite{DBLP:Lepikhin2021gshard} & \multicolumn{1}{c|}{$968.1 \pm 1.4$} & \(707.7 \pm 1.6 (1.37\times) \)\\ \hline
X-MoE~\cite{chi2022representation} & \multicolumn{1}{c|}{\(1064.0 \pm 1.5\)} & \(746.9 \pm 2.8 (1.42\times) \) \\ \hline
Sigmoid~\cite{lewis2021base} & \multicolumn{1}{c|}{\(986.6 \pm 1.4\)} & \(721.0 \pm 1.8 (1.37\times)\) \\ \hline
EC~\cite{zhou2022mixture} & \multicolumn{1}{c|}{\(909.9 \pm 1.8\)} & \(685.5 \pm 1.5 (1.33\times)\) \\ \hline
\end{tabular}
\label{tab:more_gate}
\end{table}

\section{Related Work}
In optimizing the training performance of MoE models, there are three main orthogonal directions that have been explored. These directions include MoE algorithms, AlltoAll algorithms, and scheduling algorithms. While MoE algorithms focus on workload balancing and designing gating functions, and AlltoAll algorithms aim to improve data dispatch and combine efficiency, our primary focus lies on MoE systems and scheduling algorithms that aim to reduce communication time, so we mainly introduce the related studies in this direction.

Tutel~\cite{hwang2023tutel} and DeepSpeed-MoE~\cite{DBLP:Rajbhandari22deepspeedmoe} stand out as specialized optimized systems for training MoE models. These frameworks incorporate a multitude of optimization techniques. However, their current capabilities are limited to manual configuration of the pipeline degree or heuristic search methods within a constrained search space. Contrasting Tutel, FasterMoE~\cite{he2022fastermoe} allows partitioning input tokens into two groups for the overlaps between expert computations and AlltoAll communications. Built on Tutel, PipeMoE~\cite{shi2023pipemoe} proposes an innovative and optimal partitioning methodology for input tokens. Lina~\cite{li2023lina} aims to alleviate network contention during backpropagation by addressing the challenges associated with AllReduce and AlltoAll operations. 

\change{Subsequently, various studies concern the fine-grain overlap between communication and computation. T3~\cite{T3} introduces a hardware-software co-design approach to seamlessly integrate serialized communication with computation, thus reducing resource conflicts. Wang et al.~\cite{overlap2023wang} enhance overlapping by using semantically equivalent graph transformations, implemented in XLA. Punniyamurthy et al.~\cite{punniyamurthy2024optimizingdistributedmlcommunication} tackle the issue of collective communication overhead in DLRM. FLUX~\cite{flux} and CoCoNet~\cite{coconet} break down the initial communication and computation into much smaller, more detailed tiles compared to current methods. Subsequently, it combines the tiled computation and communication into a unified kernel. Shi et al.~\cite{shi2021exploiting} propose to exploit simultaneous communication streams to improve the bandwidth utilization of AllReduce communications. Their approaches could enhance our method by addressing the competition for resources between communication and computation.}




\section{Conclusion}
In this work, we present a flexible training system
named \modelname{} to optimize task scheduling. To achieve this goal: 1) we design unified abstraction and online profiling of MoE modules across various MoE implementations, 2) we co-schedule intra-node and inter-node communications with computations to minimize communication overhead, and 3) we design an adaptive gradient partitioning method for gradient aggregation and a schedule to adaptively pipeline communications and computations. Experimental results on two clusters up to 48 GPUs show that our \modelname{} outperforms the state-of-the-art MoE training systems (DeepSpeed-MoE and Tutel) with speedups of $1.18\times$ to $1.22\times$ on 1458 customized MoE layers and $1.19\times$ to $3.01\times$ on real-world MoE models based on GPT-2 and Mixtral.

\section*{Acknowledgments}

We extend our heartfelt gratitude to the anonymous reviewers whose insightful and constructive feedback has been instrumental in elevating the quality of this paper. Their astute comments and suggestions have significantly contributed to refining our research work. The research was supported in part by National Science Foundation of China (NSFC) grants under Grant No. 62272122, and Grant No. 62302123, Guangdong Provincial Key Laboratory of Novel Security Intelligence Technologies under Grant 2022B1212010005, the Guangzhou Municipal Joint Funding Project with Universities and Enterprises under Grant No. 2024A03J0616, Shenzhen Science and Technology Program under Grant No. KJZD20230923115113026 and KJZD20230923114213027, a RGC RIF grant under the contract R6021-20, RGC TRS grant under the contract T43-513/23N-2, a Hong Kong RIF grant under the Grant No. R6021-20, Hong Kong CRF grants under Grant No. C2004-21G, C7004-22G, C1029-22G, and C6015-23G, and RGC GRF grants under the contracts 16200221, 16207922 and 16207423. Shaohuai Shi and Xiaowen Chu are the corresponding authors.


\bibliographystyle{plain}
\afterpage{\afterpage{\afterpage{\enlargethispage{-6cm}}}}
\bibliography{cite}
\end{document}